%% file: main.tex
\documentclass[acmsmall]{acmart}
\setcopyright{none}
\usepackage[vlined,ruled]{algorithm2e}
\usepackage{algorithmic}
\usepackage[inline]{enumitem}
\usepackage{graphicx} 
\usepackage{makecell}
\usepackage{colortbl}
\usepackage{mdframed}
\usepackage{mathtools}
\usepackage{wrapfig}
\usepackage{alltt}
\usepackage{xspace}
\usepackage{subcaption}
\usepackage{tikz-qtree}
\usepackage[inline]{enumitem}

\begin{document}

\input{commands}
\input{colors}

\title[Marriage of Pre-trained Language Models and Component-based Synthesis]{\papername}

\author{Kia Rahmani}
\orcid{0000-0001-9064-0797}
\affiliation{
  \position{Research Assistant}
  \department{Department of Computer Science}              
  \institution{Purdue University}            
  \city{West Lafayette}
  \state{Indiana}
  \country{USA}                    
}
\email{rahmank@purdue.edu}          
\additionalaffiliation{The first author worked on this paper during an internship with the PROSE team at Microsoft}

\author{Mohammad Raza}
\affiliation{
  \institution{Microsoft}            
  \country{USA}                    
}
\email{moraza@microsoft.com}

\author{Sumit Gulwani}
\affiliation{
  \institution{Microsoft}            
  \country{USA}                    
}
\email{sumitg@microsoft.com}

\author{Vu Le}
\affiliation{
  \institution{Microsoft}            
  \country{USA}                    
}
\email{levu@microsoft.com}

\author{Daniel Morris}
\affiliation{
  \institution{Microsoft}            
  \country{USA}                    
}
\email{Daniel.Morris@microsoft.com}

\author{Arjun Radhakrishna}
\affiliation{
  \institution{Microsoft}            
  \country{USA}                    
}
\email{arradha@microsoft.com}

\author{Gustavo Soares}
\affiliation{
  \institution{Microsoft}            
  \country{USA}                    
}
\email{Gustavo.Soares@microsoft.com}

\author{Ashish Tiwari}
\affiliation{
  \institution{Microsoft}            
  \country{USA}                    
}
\email{astiwar@microsoft.com}

\begin{abstract}
Multi-modal program synthesis refers to the task of synthesizing programs
(code) from their specification given in different forms, such as a
combination of natural language and examples. Examples provide a precise but
incomplete specification, and natural language provides an ambiguous but
more ``complete'' task description.
Machine-learned pre-trained models (PTMs) are adept at handling ambiguous natural 
language, but struggle with generating syntactically and semantically precise code.
Program synthesis techniques can generate correct code, often even from incomplete
but precise specifications, such as examples, but they are unable to work with
the ambiguity of natural languages.  We present an approach that combines 
PTMs with component-based synthesis (CBS): PTMs are used to generate candidates
programs from the natural language description of the task, which are then used
to guide the CBS procedure to find the program that matches the precise
examples-based specification.
We use our combination approach to instantiate multi-modal synthesis systems for two programming domains:
the domain of regular expressions and 
the domain of CSS selectors. Our evaluation demonstrates the effectiveness of our domain-agnostic approach in comparison to a state-of-the-art specialized system, and the generality of our approach in providing multi-modal program synthesis from natural language and examples in different programming domains.

\end{abstract}

\maketitle

\input{intro}

\input{prog}

\input{synth}
\input{opt_new}

\input{eval}
\input{related}

\input{future}

\input{conclusion}

\bibliography{ref}

\appendix
\input{app}

\end{document}

%% file: commands.tex
\newcommand{\tool}[0]{\textsc{tool}\xspace}        
\newcommand{\papername}[0]{Multi-modal Program Inference: a Marriage of Pre-trained Language Models and Component-based Synthesis} 
\newcommand{\shorttitle}[0]{Multi-modal Program Inference}

\newcommand\ignore[1]{}

 \newcommand\COM[1]{\textcolor{red}{[\texttt{#1}]}}  
 \newcommand\REV[1]{\textcolor{black}{{#1}}}  
\newcommand\arsays[1]{\COM{AR: #1}}

\newcommand{\intt}[0]{\mathbb{Z}}   
\newcommand{\natt}[0]{\mathbb{N}}   
\newcommand{\booll}[0]{\mathbb{B}} 
\newcommand{\ALT}{\,\mid\,}
\newcommand{\reg}[1]{\textcolor{reg}{\mathsf{#1}}}
\newcommand{\css}[1]{\textcolor{css}{\mathsf{#1}}}
\newcommand{\lang}[0]{\mathcal{L}}  
\newcommand{\model}[0]{\mathcal{M}}  
\newcommand{\natl}[0]{N}  
\newcommand{\wt}[0]{\mathsf{WT}}  
\newcommand{\cl}[0]{\mathsf{CL}}  
\newcommand{\inp}[0]{\mathtt{in}}  
\newcommand{\outp}[0]{\mathtt{out}}  
\newcommand{\regex}[0]{\mathtt{REGEX}}
\newcommand{\cache}[0]{\mathbb{C}}
\newcommand{\ex}[0]{E}
\newcommand{\seq}[1]{\widetilde{#1}}
\newcommand{\opcnt}[0]{\mathtt{cnt}}

\newcommand{\set}[1]{\overline{#1}}
\newcommand{\func}[1]{\textcolor{param}{\textsc{#1}}}
\newcommand{\denote}[1]{{[\![#1]\!]}}
\makeatletter
\newcommand{\customline}[1]{%
  \let\old@ALC@lno=\ALC@lno%
  \renewcommand{\ALC@lno} { \Small \textcolor{grey5} {#1\old@ALC@lno}
    \let\ALC@lno=\old@ALC@lno}%
}
\makeatother

\newcommand{\sort}[0]{\mathtt{Sort}}  
\newcommand{\op}[0]{\mathtt{Oper}}  
\newcommand{\const}[0]{\mathtt{Const}}  
\newcommand{\sigret}[0]{\psi_\mathtt{ret}} 
\newcommand{\sigin}[0]{\psi_\mathtt{arg}} 

\newcommand{\domin}[0]{\Delta_\mathtt{in}}  
\newcommand{\domout}[0]{\Delta_\mathtt{out}}  

\newlength{\emstr}
\setlength{\emstr}{0.32em plus 1ex minus 1ex}
\newcommand{\ourpara}[1]{%
\vspace*{3pt}%
\par\noindent\textbf{\textit{#1}}\hspace{\emstr}
}%

\def\multiset#1{\ensuremath{\{\kern-.33em\{#1\}\kern-.33em\}}}

\newtheorem{ass}{Assumption}[section]

%% file: colors.tex
\definecolor{reg}{RGB}{130, 10, 10}
\definecolor{css}{RGB}{130, 10, 10}
\definecolor{tick}{RGB}{0, 120, 0}
\definecolor{eff}{RGB}{10, 10, 170}
\definecolor{param}{RGB}{10, 70, 10}

\definecolor{grey12}{rgb}{0.98,0.98,0.98}
\definecolor{grey11}{rgb}{0.96,0.96,0.96}
\definecolor{grey10}{rgb}{0.93,0.93,0.93}
\definecolor{grey9}{rgb}{0.9,0.9,0.9}
\definecolor{grey8}{rgb}{0.8,0.8,0.8}
\definecolor{grey7}{rgb}{0.7,0.7,0.7}
\definecolor{grey6}{rgb}{0.6,0.6,0.6}
\definecolor{grey5}{rgb}{0.5,0.5,0.5}
\definecolor{grey4}{rgb}{0.4,0.4,0.4}
\definecolor{grey3}{rgb}{0.3,0.3,0.3}
\definecolor{grey2}{rgb}{0.2,0.2,0.2}

%% file: intro.tex
\section{Introduction}

In recent years, pre-trained language models (PTMs) have made major breakthroughs in natural language understanding. Models such as Google’s BERT  \cite{bert} and OpenAI's \mbox{GPT-3}~\cite{brown2020language} demonstrate the potential for \emph{artificial general intelligence} (AGI), in how they provide a powerful basis for creating robust natural language applications without the need for significant domain-specific training.
In particular, GPT-3 (generative pre-trained transformer) is a powerful model that can be viewed as an intelligent conversation completion engine: 
given some text in a so-called {\em{prompt}}, the model predicts the ``most sensible'' text that can follow that prompt. The predicted text tries to maintain the flow of the text in the prompt. 

GPT-3 has generated a lot of excitement by enabling a wide variety of tasks through {\em{few-shot learning}}~\cite{gpt3apps}.
 Few-shot learning refers to the fact that the completion predicted by the model can be tuned by providing only a handful of completion examples in the prompt.
For example, if the prompt contains some examples of natural language (NL) sentences being followed by the sentiment they convey, and the prompt ends with a sentence, then GPT-3 will predict the sentiment for that last sentence.
It is able to do this surprisingly well because it has been trained on the huge amounts of data available on the web.

It is natural to wonder (as many indeed have in various internet discussions and blog posts) if 
few-shot learning with PTMs can be used to go from NL descriptions to code.
For example, if we craft the prompt to include some examples of natural language descriptions followed by code, and then we end the prompt with a natural language description, GPT-3 will predict code that best completes the conversation presented in the prompt.
Does that mean that GPT-3 has solved the challenge of generating code from natural language descriptions?

While the generality of PTMs is extremely powerful, it usually comes at the cost of limited precision.
We observed that PTMs frequently fail to find exactly the right program from the given NL description, though they may output programs that are very similar to the correct one. We can also configure the PTM to return multiple programs, which it samples from some probability distribution over programs implied by the NL description, but this set also commonly does not contain the desired program due to the many possible variations. As natural language is ambiguous and imprecise, in many cases it is just not possible (even for a human) to infer the precise intent from a natural language description alone. This motivates the need for
so-called \emph{multi-modal interaction} paradigms \cite{regel,manshadi2013, raza2015}, where the user can provide a natural language description together with  specific input-output examples to precisely express their intent for how the desired program should behave. Such multi-modal interaction is also natural in human interactions as observed in help forum discussions where users convey their intent with a mixture of natural language descriptions and concrete examples.

If  we
 are given examples in addition to the natural language description, the main question that arises is how the examples can be leveraged to improve the results produced by a language model such as a PTM. What we observe is that although the PTM's candidate programs often do not contain precisely the correct program, the programs in this set often contain many relevant {\emph{components}} (sub-expressions) and use the relevant {\emph{operators}} -- but that these are just not composed correctly to produce the right program.
This leads to the idea that the set of candidate programs produced by a PTM can be effectively leveraged by a \emph{component-based program synthesis} technique to construct the desired program from a multi-modal task specification.
Component-based synthesis (CBS) \cite{SyGus, eusolver,gulwani2011loopfree,Feng2017} is a generic approach for synthesizing a program in a domain-specific programming language (DSL) that satisfies a given formal specification of a task (such as input-output examples). In its simplest form, CBS is a systematic enumerative search in the space of possible programs defined by the DSL. It begins with a set of components that are well-formed expressions in the DSL, and iteratively constructs larger programs by combining these components using the operators of the DSL, until a program that satisfies the specification is found. However, in practice, the main challenges for any CBS technique is to handle the state space explosion due to the exponential growth of the set of possible programs, as well as the challenge of ranking among many possible synthesized programs that may satisfy the given specification.

In this work we address these challenges by introducing a generic approach to multi-modal program inference that  is based on  a marriage of pre-trained natural language models and component-based program synthesis. Our approach combines the benefits of the two techniques by leveraging the output of a PTM to guide all three key phases of the CBS search: the initialization of components, the iterative synthesis of larger programs, and the ranking of final candidate programs. In this way, the combination of the two approaches serves to address the short-comings of each: the CBS synthesis improves precision by constructing a program that satisfies the examples (which may otherwise not appear in any of the PTM outputs), while the PTM output tames the complexity of the CBS search space by guiding the search at every stage.  

A notable characteristic of our approach is its generality that comes from its \emph{domain-agnostic} design: it has not been designed for any particular domain-specific language and can in principle be applicable to different programming domains. As our primary domain of study we focus on the language of regular expressions and illustrate the benefits of our approach in comparison to the state-of-the-art for multi-modal synthesis techniques designed especially for this domain. We also illustrate the generality of our approach by presenting a concrete instantiation and evaluation in the very different domain of CSS selectors, which is a DSL used in web programming. 
\REV{Note that we do not claim that our approach can be directly applied to any arbitrary programming language off-the-shelf, but only that it is not limited to one particular language by showing its applicability and benefits in at least two very different programming domains. In section 7 we also discuss some limitations and expected improvements as we consider scaling to other domains, but a broader evaluation and extension of these techniques to arbitrary languages is left for future work.}

\subsection{Motivating examples and overview of approach}

\input{Figs/Motiv}

Consider a scenario where a user needs help writing a regular expression. Figure~\ref{fig:motivate} shows three such tasks. The first column shows the natural language (NL) description the user provides, and the second column shows the ground truth the user desires. (The first two tasks are from the dataset in \cite{kushman2013} and the third one is from a Stack Overflow question.)


The first step in our approach is to use a PTM to generate candidate regular expressions from the NL description.
Throughout this work, the PTM we use is Open AI's GPT-3 system \cite{brown2020language}, which is a state-of-the-art pre-trained model for code generation from natural language. To get a PTM to produce regular expressions, we need to provide it with the {\em{right}} query (called a prompt). We exploit the few-shot learning capabilities of PTM and provide it with the best possible prompt using our novel dynamic prompt generation algorithm (Section~\ref{sec:opt}), which is inspired by literature on information retrieval. 
PTMs internally generate a probability distribution on possible completions, and then they
sample
 from this distribution to generate individual candidates. We exploit this fact to configure the PTM to generate a diverse sample.
The third column in Fig~\ref{fig:motivate} 
shows
some sample candidates 
returned by the PTM.

In all three cases, the first observation we can make is that the results of the PTM in general look very similar to the ground truth, but none of them are exactly equivalent to it. This can be expected given the significant ambiguity in the NL descriptions which is difficult even for a human to resolve. For instance, for task I it is not clear if the intent is that any of "!", capital or lower-case can occur before the character, or if only the lower-case is permitted to occur before a character and the other two should occur alone on the line (the ground truth shows that the intent is the latter). It is also not clear if "before a character" should mean immediately before a single character or not. Similar ambiguities can be seen in the other two tasks, e.g. whether "at least 7" refers to just "!" or not in task II, whether "at least zero times" refers to everything before it in task III and whether "followed by" means immediately followed by or not. 

Such ambiguities are common in natural language, and a good way to resolve them is by allowing the user to provide concrete examples of the desired behaviour, such as examples of strings the intended regex should or should not match. Such multi-modal intent specification is natural even in human interactions as can be seen in help-forum questions where users often provide a description of the task as well as concrete examples to express their intent. Given such examples in addition to the NL, the challenge is how to generate the correct program. The second step in our approach addresses this challenge using a component-based synthesis (CBS) technique, which is designed to utilize the candidates provided by the PTM at each of the three key phases of the search (initialization, expansion and ranking), which we illustrate next. 

{\bf Initialization}. One important question for any CBS algorithm is how to obtain the initial set of components to begin the search. In an extreme brute force search, one may initialize with a set of concrete values for every terminal symbol of the DSL grammar (e.g. all possible character values in the regex domain), but this is practically untenable for any non-trivial DSL. In our case we observe that the candidate programs provided by the PTM all contain very relevant components that can be used to construct the correct program. For example, for case I in Figure \ref{fig:motivate} we observe frequently occurring relevant components such as "$[A\!-\!Z]$", "$[a\!-\!z]$", "$!$" and "$.*$". For case II, apart from important frequent components such as "$!$" and the number $7$, we can observe even the prominent occurrence of the large sub-expression "$[aAeEiIoOuU]$" that represents the notion of a vowel that the PTM has identified. Such an expression using many occurrences of the class union operator would require prohibitively many iterations and examples to construct if starting from purely atomic components. This leads to the question of how we can obtain these most prominent sub-expressions from the PTM outputs, which we address with the novel notion of \emph{maximal components}. Intuitively, these are the largest sub-expressions that occur in the PTM candidates with high frequency. We demonstrate how starting from such maximal components  can help to effectively construct the correct program as compared to 
the traditional component-based approach that starts from all atomic components.

{\bf Expansion}. After creating the initial set of components, the CBS approach proceeds by iteratively creating larger programs. At each iteration, this is done by applying the DSL operators to the existing programs to create larger programs. The brute force approach would be to exhaustively apply every operator on every combination of components as permitted by the type system of the DSL, but this leads to a combinatorial blowup in practice. A more tractable option is to employ a \emph{beam search} approach where only a bounded number of new programs are kept at every iteration, but the main question is what criteria to use for which programs should be kept or disregarded. 

We address this question again using the PTM candidates, by observing the frequency distribution of operators that is found in these programs and biasing the beam search with respect to this distribution. For instance, for case I in Figure \ref{fig:motivate} we observe that operators such as alternation (|) and iteration (*) are used about once or twice on average across all candidate programs, while other operators such as quantifiers or character class negation are not used at all. This signals a preference for programs that follow a similar operator distribution pattern as opposed to programs that may use five alternations. Technically, we compute an operator frequency distribution vector from the set of PTM candidates, and at each iteration of the beam search we maintain a bounded set of new programs that most closely follow this distribution. In addition, unlike standard beam search methods, we also maintain semantic variety in the beam exploration by ranking within semantic equivalence classes of programs rather than a global ranking in the search space. Such \emph{condensing} of the set of programs within equivalence classes minimizes redundant syntactic variations of the same program in the search exploration. 

{\bf Ranking}. Eventually, the goal of the CBS algorithm is to return a synthesized program to the user that satisfies the examples. But after a certain number of iterations of CBS in practice, there can be a large number of programs that satisfy the given examples. Hence the main question is how to rank among these programs. This decision can be guided by considering similarity of the synthesized programs to the PTM candidates. The operator frequency  distribution as used above is a good signal for guiding the search in terms of which operator applications to explore, and is also a good indicator for the final preference of which program to pick from the set of synthesized programs. However, we also found that for final ranking it is helpful to use additional stronger signals such as direct string similarity of programs to the PTM candidates. We found a combination of these signals more finely distinguishes between the final set of synthesized programs in terms of how different operators are being used in the program.

\subsubsection*{Contributions} The core contributions we make in this work are summarized as follows. 
\begin{itemize}
    \item We present an abstract domain-agnostic formulation of a multi-modal program inference algorithm that can synthesize programs in an arbitrary DSL when given a natural language description and examples of an intended task. This algorithm uses a novel CBS synthesis technique that utilizes the output of a PTM on the given NL description to generate a program that satisfies the given examples, and we demonstrate the relative completeness of our approach with respect to the PTM output.
    \item We present a concrete instantiation of our technique for the domain of regular expressions, that has been a popular domain in many works that have explored programming by natural language, examples and multi-modal approaches. We present an evaluation of our approach as compared to the state-of-the-art specialized technique for multi-modal regex synthesis, on both existing and new datasets. 
    \item We present secondary instantiation and evaluation of our approach in the very different domain of CSS selectors for extracting elements from web pages. \REV{This illustrates the generality of our approach and its applicability in at least two different practical programming domains.}
    \item We show how the \emph{prompt} provided to pre-trained models such as GPT-3 can significantly impact the quality of results, and present novel techniques for formulating this prompt based on ideas from information retrieval~\cite{sparckjones1972} to show how GPT-3 results can be significantly improved.
\end{itemize}

%% file: Figs/Motiv.tex
\begin{figure*}[t]
    \centering
    \scalebox{0.69}{%
    \begin{tabular}{ | c | l | l | l |}
    \hline
    \textbf{\#} &
     \multicolumn{1}{|c|}{\textbf{Natural Language}} & \multicolumn{1}{|c|}{\textbf{Ground Truth}} & 
     \multicolumn{1}{|c|}{\textbf{ Pre-trained Model's Candidates}}
      \\ \hline \hline
    \textsc{i} &
    $
    \begin{array}{l}
    \text{A line with a "!", a capital, or a }\\
    \text{lower-case before a character} \\ 
    \end{array}
    $
    &
    $(!|[A\!-\!Z])|[a\!-\!z].*..*$
    &
    $
    \arraycolsep=16pt
    \hspace{-4mm}
    \begin{array}{ll}
      {(!|.|([a\!-\!z])|([A\!-\!Z]))} \quad &
      { [A\!-\!Z].*(.|[a\!-\!z]).*}   \\
      {[A\!-\!Z]!|[a\!-\!z]!|[a\!-\!z]?} \quad &
       {([A\!-\!Z][a\!-\!z]).*(.|.|.).*}   \\
       {(!|[a\!-\!z]|[A\!-\!Z][a\!-\!z]).*} \quad &
       {!|[a\!-\!z]|[A\!-\!Z]} \\
        {([a\!-\!z]|[A\!-\!Z]).*.} \quad &  {(.|[A\!-\!Z]).*.|[a\!-\!z]} \\
    \end{array}
    $
    \\ \hline
    \textsc{ii} & 
    $
    \begin{array}{l}
      \text{Lines with at least 7 of the } \\
      \text{string "!" or a vowel }
    \end{array}
    $
    &
    ${(!|[aAeEiIoOuU])\{7,\}}$
    & 
    $
    \hspace{-0.6mm}
    \arraycolsep=6.5pt
    \begin{array}{ll}
    {((!\!*7)|[aAeEiIoOuU])\!+.*}
    \quad &
    {((!+)|[aAeEiIoOuU])+}
    \\
    {(!\!+.\!*|[aAeEiIoOuU]\{7\})}
    \quad &
    {(.*!)\{7,\}|(.\!*[aAeEiIoOuU])}
    \\
    {(.*[aAeEiIoOuU]|{vowel})\{7\}}
    \quad &
    {((!+)|.......)+}
    \\
    {((!!)\!+|[aAeEiIoOuU])+}
    \quad &
    {((((!.?)?)*)|[aAeEiIoOuU])+}
    \end{array}
    $
     \\  \hline 
    \textsc{iii} &
    $
    \begin{array}{l}
    \text{At least one digit followed by }\\
    \text{character : at most once followed } \\ 
    \text{by a digit at least zero times}
    \end{array}
    $
    &
    $[0\!-\!9]+\!:?[0\!-\!9]*$
    &
    $
     \hspace{-0.5mm}
    \arraycolsep=7.4pt
    \begin{array}{ll}
        {([0\!-\!9]\!\!*\!..\!:\!([0\!-\!9]*)?)+}   \quad  &  {([0\!-\!9]\!*([:][0\!-\!9]*))\!*\!(0[0\!-\!9]+)}  \\
        {([0\!-\!9]+:)?[0\!-\!9]?}  \quad &
        {[0\!-\!9]\{3\}}  \\
        {([0\!-\!9]?:[0\!-\!9]?)*}   \quad  & 
        {([0\!-\!9]\!*..:\!*[0\!-\!9]\!*0*)*}  \\
        {([0\!-\!9]\{1,\}(?\!:\!.[0\!-\!9]\{0,\}))*} \quad &  {([0\!-\!9]\{3\})+}  \\
    \end{array}
    $
 \\
 \hline
    \end{tabular}
    }
    \vspace{0mm}
    \caption{Examples of three tasks with natural language descriptions of regular expressions, the intended ground truth program, and a sample of the pre-trained model's top-ranked candidates for the task}
    \label{fig:motivate}
\end{figure*}

%% file: prog.tex
\section{Domain Specific Languages}
\label{sec:dsl}

Our multi-modal program synthesis algorithm is not designed for a particular programming domain and is  parameterized by an arbitrary domain-specific language (DSL) and its execution semantics. In this section, we formally define the general notion of DSLs we use. We also illustrate this by introducing the language of regular expressions, which is the main DSL studied in this paper, and the language of CSS selectors which we also study as a secondary domain. 

A DSL is defined as a tuple $\lang:=(\sort,\const,\op,\mathtt{s}^\circ,\sigin,\sigret)$ where 
$\sort$ is a set of \emph{sorts}, 
$\const$ is a set of \emph{constants},
$\op$ is a set of \emph{operators}, and 
$\sigin:\op\rightarrow\seq{\sort}$
and 
$\sigret:(\op\cup\const)\rightarrow\sort$ are a pair of  \emph{signature} functions. The signature function
$\sigin$ maps a given operator to an ordered sequence of sorts -- \emph{of its arguments} --
and the signature function
$\sigret$ maps a given operator or a constant to a single sort -- \emph{of its  return value}. 
A DSL can be used to build {\em{terms}} as follows: every constant is a term, and if
$t_1, \ldots, t_n$ are terms of sorts $s_1,\ldots, s_n$ respectively, then
$\mathsf{op}(t_1,\ldots,t_n)$ is a term of sort $\sigret(\mathsf{op})$ if
$\sigin(\mathsf{op}) = \langle s_1,\ldots,s_n\rangle$. We do not distinguish between a DSL and the set of terms it generates, and hence,
$\lang$ will also denote the set of all terms in the language. 
Each DSL also contains a special sort $\mathtt{s}^\circ$  and the goal of the synthesis algorithm is to return a term of this sort. 
Terms of sort $\mathtt{s}^\circ$
will be called \emph{closed} terms or \emph{complete programs}.

\tikzset{sibling distance=2pt}
\tikzset{level distance=19pt}
\begin{figure}[t]
\begin{subfigure}[b]{.46\textwidth}
\begin{footnotesize}
 \scalebox{1.05}{%
$
\begin{array}{lcl}
\reg{i} & := & \{\reg{0,1,2,3,\dots}\}\\
\reg{c} & := & \{\reg{A,B,\dots,a,b,\dots, \#,\$,\%,\dots, 0,1,2,3,\dots}\}\\
\reg{s} & := & \reg{fromChar(c)}\ALT \reg{range(c,c)} \ALT \reg{union(s,s)} \ALT \\
        &    & \reg{negate(s)} \ALT \reg{any()} \\
\reg{e} & := & \reg{quant(e,i,i)} \ALT \reg{quantMin(e,i)} \ALT \reg{alter(e,e)} \ALT \\
        &    & \reg{concat(e,e)} \ALT \reg{fromCharSet(s)} \\
\end{array}
$
}
\end{footnotesize}
\caption{}
\end{subfigure}
\hfill  
\begin{subfigure}[b]{.45\textwidth}
\centering
 \scalebox{1.05}{%
\begin{tikzpicture}[scale=.75]
\Tree [.$\reg{concat}$ [.$\reg{quantMin}$ [.$\reg{fromCharSet}$ [.$\reg{range}$ $\reg{0}$ $\reg{9}$ ] ]  $\reg{1}$ ]  [.$\reg{concat}$ [.$\reg{quant}$ [.$\reg{fromCharSet}$ [.$\reg{fromChar}$ $\reg{:}$ ] ] $\reg{0}$ $\reg{1}$ ]  [.$\reg{quantMin}$ [.$\reg{fromCharSet}$ [.$\reg{range}$ $\reg{0}$ $\reg{9}$ ] ]  $\reg{0}$ ] ] ]
\end{tikzpicture}
}
\caption{}
\end{subfigure}
\vspace{-2mm}
  \caption{The DSL $\lang_{\textsc{reg}}$ of regular expressions (left) and the parsed tree of $[0\!-\!9]+\!:?[0\!-\!9]*$ \,(right)} 
  \label{fig:regex}
\end{figure}
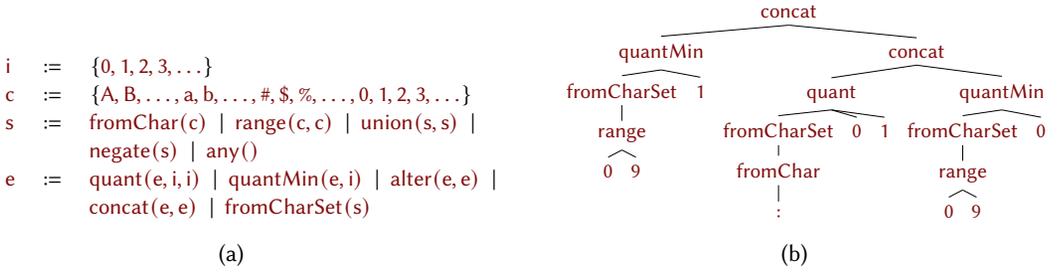

As a concrete example,
 consider \autoref{fig:regex}(a) which presents, $\lang_{\textsc{reg}}$, the DSL of regular expressions. This DSL contains four sorts, $\sort := \{\reg{i}, \reg{c}, \reg{s}, \reg{e}\}$, which respectively represent $\reg{i}$ntegers, $\reg{c}$haracters, character $\reg{s}$ets, and $\reg{e}$xpression sorts. Terms of sort $\reg{e}$ are closed.
 The set of constants $\const$ includes all non-negative integers (with sort $\reg{i}$) and all characters (with sort $\reg{c}$).
 There are also ten operators in the set $\op$ of operators, whose signatures are shown
in \autoref{fig:regex}(a).
For instance, $\reg{quant}$ is an operator with 
$\sigin(\reg{quant})\!=\!\langle\reg{e},\reg{i},\reg{i}\rangle$
and 
$\sigret(\reg{quant})\!=\!\reg{e}$. This DSL encodes a large set of regular expressions that developers commonly write; for example, \autoref{fig:regex}(b) presents the parsed syntax tree of the ground truth expression in \autoref{fig:motivate} (\#\textsc{iii}). 

\begin{definition}[sub-term and atomic terms]
\label{def:subterm}
A reflexive and transitive \emph{sub-term} (or \emph{sub-component}) relation, denoted by $\sqsubseteq: \lang\times\lang$,
holds between terms $t$ and $t'$, denoted by $t \sqsubseteq t'$, if $t$ appears as an argument in the syntax tree of $t'$. 
We say that a term $t$ is \emph{atomic} if there does not exist any other term $t'$ such that  $t' \neq t$ and $t' \sqsubseteq t$. 
\end{definition}

For example, 
$\reg{fromCharSet(range(0,9))}$  is a sub-term of the expression shown in \autoref{fig:regex}(b), and terms $0$ and $9$ are atomic.

We now formulate a general notion of semantics for terms. 
We assume that there is an input domain $\domin$
and an output domain $\domout$ for each sort, and the
semantics of a DSL is specified by a function 
$\denote{.}:\lang\rightarrow(\domin\rightarrow \domout)$ 
that given a term in $\lang$ returns a function from the input domain to the output domain. Under these assumptions, terms can be viewed as \emph{programs} which transform an input from $\domin$ to an output in $\domout$. We may also refer to any sub-term of a complete program as a \emph{component} of that program.%

For instance, the semantics of closed terms in the DSL of regular expressions is defined over the input domain $\domin$ that contains all finite strings and the output domain $\domout := \{\bot,\top\}$. A string $str$ is said to be \emph{accepted} by the expression $r$ if and only if $\denote{r}(str) = \top$. We adopt this semantics from the standard regular expression implementations. Informally\footnote{The formal definition of this semantics is provided in \autoref{app:def}.}, a term of sort $\reg{s}$ accepts strings containing a single character $c$ if and only if $c$ satisfies constraints imposed by the root operator of that term. In particular, $\reg{fromChar}(c_1)$ only accepts the character $c_1$, $\reg{range}(c_1,c_2)$ accepts any character that lies between $c_1$ and $c_2$, $\reg{union}(s_1,s_2)$ accepts characters that are accepted by either $s_1$ or $s_2$ and $\reg{negate}(s)$ accepts characters which are \emph{not} accepted by $s$. The $0$-ary operator $\reg{any()}$ accepts all characters. 

In a similar fashion, the operator $\reg{quant}(e,i,j)$ only accepts strings composed of $k$ sub-strings (for any $i\!\leq\! k\!\leq\! j$) each of which is accepted by $e$, and operator $\reg{quantMin}(e,i)$ is semantically equivalent to $\reg{quant}(e,i,\infty)$. Operator $\reg{alter}(e_1,e_2)$ accepts strings accepted by either $e_1$ or $e_2$ and operator $\reg{concat}(e_1,e_2)$ only accepts strings of the form $str_1;str_2$ if $e_1$ accepts $str_1$ and $e_2$ accepts $str_2$.
Operator $\reg{fromCharSet}$ does not impose any restriction on the accepted strings and simply \emph{lifts} the terms from sort $\reg{s}$ to the closed sort $\reg{e}$.

For example, the closed term $\reg{fromCharSet(range(0,9))}$ accepts any string composed of a single digit and the term 
presented in \autoref{fig:regex}(b) accepts all the following four bold-faced strings: 
\textbf{1991:10}, \textbf{99999}, \textbf{0:1} and \textbf{000:}.

As a second target, we now present the domain of 
Cascading Style Sheets (CSS) selectors.  Figure~\ref{fig:cssdsl} shows, $\lang_{\textsc{css}}$, the DSL for CSS selectors. CSS selectors are
expressions for selecting elements from the document object model (DOM) of a webpage. 
They select nodes based on structural properties that are defined by the HTML source markup of the webpage.
For instance, the CSS selector $\css{AttributeEquals(TagEquals(Any(),}$
$\css{"div"),}$
$\css{"class",}$ $\css{"row")}$, call it $css_1$, selects {\em{all nodes with tag "div" and class "row"}}, which is typically written as $\css{div.row}$. 
Similarly, the CSS selector $\css{Children(css_1, AttributeEquals(Any(), "id", "myid"))}$ picks {\em{all nodes that have id "myid" that are immediate child of any node with tag "div" and class "row"}}, which is typically written as $\css{div.row > \#myid}$,
and the CSS selector 
$\css{AttributeEquals(}$
$\css{nthChild(}$
$\css{TagEquals(}$
$\css{Any(), "li"),}$ 
$\css{MultipleOffset(2,0)),}$
$\css{"hidden",}$ 
$\css{"true")}$ 
represents {\em{all nodes with tag "li" whose attribute "hidden" is set to "true" and that occurs at even positions in the sibling list}}, which is typically written as $\css{li}\css{:}\css{nth\mbox{-}child(2n)[hidden="true"]}$. The formal semantics is provided in Appendix~\ref{app:def}. CSS selectors are needed when scraping data from web, or when doing web programming in general. They can be hard to write manually, especially for an occasional user, but they are often easy to describe in natural language.

Having defined the syntax and the semantics of domain specific languages 
and in particular the DSL of regular expressions and CSS selectors,
in the next section, we will formally introduce multi-modal synthesis tasks and 
describe in detail our generic CBS solution for those tasks.

\begin{figure}[t]
\begin{footnotesize}
 \scalebox{1.05}{%
$
\begin{array}{lcl}
\css{s} & := & \css{"a\ string\ literal"} \\
\css{i} & := & \css{a\ number\ literal} \ALT \css{MultipleOffset(i,i)} \\
\css{n} & := & \css{Any()} \ALT \css{Union(n,n)} \ALT \css{Not(n,n)} \ALT \css{TagEquals(n,s)} \ALT \css{nthChild(n,i)} \\
        &    & \css{AttributeEquals(n,s,s)} \ALT \css{nthLastChild(n,i)} \ALT \css{AttributeContains(n,s,s)} \ALT \css{RightSibling(n,n)} \\
        &    & \css{AttributeStartsWith(n,s,s)} \ALT \css{Children(n,n)} \ALT \css{AttributeEndsWith(n,s,s)} \ALT \css{Descendants(n,n)} 
\end{array}
$
}
\end{footnotesize}
    \caption{The DSL $\lang_{\textsc{css}}$ of CSS expressions.} 

\label{fig:cssdsl}
\end{figure}

%% file: synth.tex
\section{Multi-modal Program Synthesis Algorithm}
\label{sec:synth}

\input{Figs/algorithm}

In this section we present our multi-modal program synthesis algorithm that synthesizes a program to accomplish a task specified in terms of natural language and examples. Our algorithm is \emph{domain-agnostic} and is parameterized by a DSL. Given a DSL $\lang:=(\sort,\const,\op,\mathtt{s}^\circ,\sigin,\sigret)$, 
we define a multi-modal synthesis task as a tuple $(\natl,\ex)$, where $N$ is the natural language description of the task and $\ex$ is a set of \emph{examples}.
We define an example $e\in\domin\times\domout$ as a pair of values from the input and the output domains. The synthesizer's goal is to find a program $p\in\lang$ that is \emph{consistent} with the given examples, defined as follows:
$$p\models\ex \Leftrightarrow \forall_{<i,o>\in\ex}\denote{p}(i)=o$$

Our algorithm, \textsc{nlx}, for multi-modal synthesis from natural language and examples is presented in \autoref{fig:alg}. The main top-level function $\func{synthesize}$ (\autoref{subfig:synthesize}) returns a program synthesized from a multi-modal task specification. As the algorithm is domain-agnostic, this function is parameterized by a DSL $\lang$ and a PTM $\model$ for this domain. In Section \ref{sec:opt} we describe the details of the particular PTM model we use and how it is configured with few-shot learning for a particular domain, and in this section we assume that such a model $\model$ is given.

We first describe the high-level structure of our \textsc{nlx} algorithm, before describing the key phases in more detail. The  algorithm proceeds by first obtaining the top-ranked programs from the PTM for the given natural language query. It then implements a component-based synthesis (CBS) that utilizes the PTM output to guide the CBS search at each of the three key phases: the initialization of components, the iterative expansion to larger programs, and the final ranking of programs. 

The $\func{synthesize}$ function implements this high-level structure of the algorithm. It initially executes the PTM on the given  description $\natl$ and stores the resulting programs in $P$ (line 3), which is used in each of the subsequent phases. 
The algorithm uses a \emph{cache} object $\cache$ to maintain the set of synthesized programs according to their sort in the DSL. A cache, denoted by
$\cache:\sort\rightarrow\set{\lang}$, is defined as a map from sorts to sets of terms in $\lang$. The cache is initialized by extracting components from the PTM candidates $P$. This initialization phase is defined by the function \func{initialize} (line 4), which we describe in more detail in §\ref{subsec:init}. Next, we enter the expansion phase in the main loop of the algorithm at line 5. At each iteration, the cache is updated with new programs synthesized by applying operators of $\lang$ on existing components in the cache. As exploring all possible operator and component combinations is intractable in practice, we employ a beam search where such combination choices are guided by the PTM candidates $P$. This is defined by the $\func{expand}$ function that is described in  §\ref{subsec:expand}. This process is repeated up to a tunable constant  $\mathtt{SynthDepth}$. Finally, the algorithm identifies closed programs in the cache which are consistent with the given examples (line 7) and then performs a ranking to choose the best program to return out of many possible ones. This ranking is based on similarity to the PTM candidates $P$ as defined by the \func{rank} function which we describe in §\ref{subsec:rank}.

We next describe each of the three key phases of the algorithm that are formally defined by the functions in \autoref{subfig:synthesize}, and end this section with a discussion of the relative completeness of our algorithm.

\subsection{Initialization of components}
\label{subsec:init}


The first step in the algorithm is to obtain the set of initial components from which to begin the search. As we have  discussed, the set of top candidate programs $P$ provided by the PTM contains very relevant components for constructing the correct program, but initializing with a very large set of components can lead to the search complexity becoming intractable. Hence we approach this question with respect to two aspects: how likely a component is to \emph{occur} in the desired program and how likely is it that a component is \emph{redundant} in the initial component set (in the sense that it can already be included as part of another larger component). Both of these questions are addressed using a probabilistic formulation with respect to the distribution of components in the PTM candidates. 

The \func{initialize} function in \autoref{subfig:init}
takes as input the set of PTM candidate programs and returns an initialized cache. This function initially extracts the set of all sub-terms of all programs in $P$ and stores it in a variable $v_1$ (line 2). 

\textbf{Component occurrence.} At line 3 we compute the probability of occurrence of each component and keep those above a tunable minimum probability threshold defined by a constant $\mathtt{PrOcc}$. The occurrence probability for a term $t$ is computed as $\mathtt{cnt}(t,P) / |P|$, which is the proportion of programs in $P$ that contain the component $t$. For example, in \autoref{fig:motivate} (\#\textsc{ii}), if we consider the set $P$ to be the 8 candidate PTM programs, and the term $t =\reg{quantMin(fromCharSet(any()),0)}$ (printed as $.*$) appears in five of the programs in $P$, then we have the occurrence probability of $t$ given by  $\mathtt{cnt}(t,P)/|P|= 5/8 = 0.625$.

The occurrence probability check ensures that terms that appear more often in the PTM's output have a higher chance of being included in the initial cache, as often times there is noise in the PTM output that includes irrelevant components that occur very infrequently. For example, in 
\autoref{fig:motivate} (\#\textsc{ii}), the term printed as $vowel$, which is clearly 
due to PTM's confusion about the task, only appears once in the candidate programs. Such terms can be easily eliminated from the initial cache by setting the occurrence probability threshold $\mathtt{PrOcc}$ appropriately. In practice, we found that a value of $\mathtt{PrOcc} = 0.1$ worked well in all our evaluations (with usually at  least 20 PTM candidates in total).  

\textbf{Component redundancy.} The second aspect we consider in the initialization of components is that of \emph{redundant components}. This is important because while many components may occur frequently, many of these may not be useful to include in the initial cache as they may already occur as sub-components of other components. With respect to our PTM candidates, if we find that a component $t$ always appears as a sub-component of another component $t'$ in all of the PTM candidate programs, then that is a strong signal that $t$ is a redundant component as it can already be included as part of  $t'$ in the cache. For example, in \autoref{fig:motivate} (\#\textsc{ii}), the term $t_1 = \reg{fromChar(a)}$ \textit{always} appears as a sub-component of the same term $t_2$ that unions all vowels (printed as $[aAeEiIoOuU]$). Similarly all terms representing subsets of vowel characters always occur only as sub-components of $t_2$ and can be considered redundant to include by themselves. The term $t_2$ however, occurs as part of many different components and is important to include as a component by itself.  In the same example, the term $t_3:=\reg{fromCharSet(fromChar(!))}$ (printed as $!$)  appears in multiple different super-terms, e.g. in $t_4:=\reg{quantMin(}t_3\reg{,0)}$ (printed as $!*$), $t_5:=\reg{quantMin(}t_3\reg{,1)}$ (printed as $!+$) and $t_6:=\reg{concat(}t_3,t_3\reg{)}$ (printed as $!!$). We note that the inclusion of both $t_2$ and  $t_3$ in the initial cache is important to construct the ground truth program in this case, since none of $t_4$, $t_5$ or $t_6$ directly appear in the ground truth, i.e. ${(!|[aAeEiIoOuU])\{7,\}}$.

Formally, at lines 6-9 in the algorithm, we compute the probability of redundancy of a component $t$ as the proportion of super-components of $t$ that occur as many times as $t$ in the PTM candidates (note that by definition no super-component can occur more times than any of its sub-components). If the redundancy probability is below a certain maximum threshold given by $\mathtt{PrRed}$, then the component is included in the initialization.

Though in general the algorithm permits the redundancy threshold $\mathtt{PrRed}$ as a tunable constant, we note that the extreme case of  $\mathtt{PrRed} = 0$ identifies a special case of \emph{maximal components} that work well in practice. These are components that occur more frequently than any of their super-components, and hence represent the PTM's identification of a component that it uses in different ways across different candidate programs, such as the vowel component in \autoref{fig:motivate} (\#\textsc{ii}). This suggests the PTM's high confidence that the component is useful but lower confidence on \emph{how} it should be used in the final program, and hence makes it a good candidate to include in the CBS search which explores many more combinations for synthesis. 

We note that this notion of maximality is not just with respect to size but both size and frequency. Hence components $t$ and $t'$ may both be maximal even if $t \sqsubseteq t'$, if $t$ occurs more frequently than $t'$. Both would be useful to consider as the PTM candidates indicate that $t$ may be used in other ways outside of $t'$.  

\textbf{Standard components.} Finally, the algorithm permits a fixed set of standard components for the DSL that should always be included (line 10). These may be any terminal or commonly-used special values for the different sorts in the language. For instance, for the regex domain we include the standard components that are the integer values 0,1 and the specially named character classes $\mathtt{\backslash d}$,$\mathtt{\backslash s}$,$\mathtt{\backslash w}$, representing digits, space and word characters. For the CSS domain, we include the any-element selector $\css{Any()}$, the integer value 1 and the empty string attribute value.

\subsection{Expansion }
\label{subsec:expand}

In this section we describe the expansion phase of our algorithm, where larger programs are iteratively constructed using the initial components and already synthesized programs.  The brute force approach would be to exhaustively apply every operator on every combination of components as permitted by the rules of  the DSL, but this leads to intractable complexity in practice. Hence the technique we use is to employ a form of \emph{beam search} where only a bounded number of new programs are kept at every iteration. This beam search is defined by the technique of \emph{pruning} the cache which determines which synthesized programs to keep and which to discard. We design this technique based primarily on the  distribution of operators that is found in the PTM candidates and biasing the beam search with respect to this distribution. We first describe the outline of the expansion phase and then describe the pruning technique in more detail in section \ref{sub:prune}

The function $\func{expand}$ is defined in \autoref{subfig:expand}, which 
given a cache $\cache$ as input returns a new cache expanded with a set of new terms based on existing terms in $\cache$. The procedure initially obtains the subset of terms in $\cache$ to be considered for expansion using a call to the $\func{prune}$ function (line 2). The procedure then iterates over all operators $op\in\op$ and constructs new terms in $\lang$ by applying $op$ on existing terms in $\cache'$ according the signature of $op$. Newly constructed terms are then stored in a variable $v_1$ (line 5). For example, assuming that the following two terms, $t_6$ and $t_7$, are in the pruned cache $\cache'$, the set of newly constructed terms, $v_1$, will include terms like $t_8:=\reg{alter(}t_6,t_7\reg{)}$ and $t_9:=\reg{concat(}t_6,t_7\reg{)}$:
\begin{align*}
    t_6 := \reg{quantMin(fromCharSet(fromChar(a)),0)} &   \text{\hspace{5mm} (printed as $a*$)} \\
    t_7 := \reg{quantMin(fromCharSet(fromChar(b)),0)}  &  \text{\hspace{5mm} (printed as $b*$)}
\end{align*}

Finally, the procedure $\func{expand}$ updates the original cache $\cache$ by adding terms in $v_1$ to $\cache(s)$, where
$s$ is the return sort of current operator $op$
(line 7). Once the loop is iterated over all operators, the procedure returns the updated cache $\cache$ as its final output (line 8).

\subsubsection{Pruning the Cache} 
\label{sub:prune}

Our technique of pruning the search space during the beam search is based primarily on the distribution of operators that is found in the PTM candidates and biasing the beam search with respect to this distribution in a way that maintains semantic variety of the synthesized programs (i.e. minimizes redundant semantically equivalent expressions in the search space). The pruning function $\func{prune}$ is defined in \autoref{subfig:prune}, which is used in the beginning of each expansion iteration to bound the number of terms considered for expansion. 

\textbf{Semantically equivalent sub-terms}. To avoid semantically redundant states, the first pruning strategy is to remove any term that is semantically equivalent to any of its sub-terms (line 4). For instance, assume that the given set of examples is $\ex_1:=\{\textbf{aa}, \textbf{ccc}\}$ and the term $t_8$ defined earlier (i.e. $\reg{alter(}t_6,t_7\reg{)}$)
is in $v_1$. Observe that $t_8$
is not semantically distinguishable from its sub-term $t_6$ with respect to $\ex_1$ (since they both accept \textbf{aa} and reject \textbf{ccc}). Consequently, any possible use-case of $t_8$ in the future iterations
can also be handled by $t_6$, and hence, $t_8$ can be eliminated from $v_1$.
This observation is formalized by defining the \textit{interpretation} 
of a term $t$ with respect to a set of examples $\ex$, denoted by $\denote{t}_\ex$,
as a set of input and output pairs, where the input belongs to an example in $\ex$ and output is generated by running $t$ on that input, i.e.
$\denote{t}_\ex:= \{<\!\!i,\denote{t}(i)\!\!> \;\ALT\; <\!\!i,\_\!\!> \in \ex \}$.
In the example discussed above, we have  $\denote{t_6}_{\ex_1} \!\!= \denote{t_8}_{\ex_1}\! = \{<\!\!\textbf{aa},\!\top\!\!>,<\!\!\textbf{ccc},\!\bot\!\!>\}$. The procedure  eliminates all terms in $v_1$ which share their interpretation (with respect to the given examples) with some of their sub-terms  (line 4). The remaining terms are stored in a fresh variable $v_2$.

\textbf{Low frequency operators.}
Our primary signal for pruning is to bias towards the structure of the PTM candidates. The first constraint we consider in this bias is to avoid DSL operators that may occur with a very low frequency (or not at all) across all of the PTM candidates. For example, 
in \autoref{fig:motivate} (\#\textsc{iii}),
the alternation operator ($\reg{alter}$) does not appear in any of the candidate programs generated by the PTM. This signals that the target program does not have many alternation operators (it has in fact none). 

We distinguish such low-frequency operators using a tunable constant $\mathtt{OpTH}$ that defines the threshold for low-frequency operators: operators that on average have fewer occurrences than this threshold are allowed at most $\mathtt{OpTH}$ occurrences. We implement this using the function $\mathtt{opVec}$ that given a term $t$ returns an integer vector composed of the number of occurrences of each DSL operator in $t$. For example, for the term $t_6$ defined above, $\mathtt{opVec}(t_6)$ is a vector that has value $1$ in the three entries assigned to $\reg{quantMin}$, $\reg{fromCharSet}$ and $\reg{fromChar}$ and has $0$ everywhere else. Using this function, we eliminate terms whose operator vector is different from programs in $P$. In particular, the procedure $\func{prune}$ eliminates terms from $v_2$ whose  \emph{Hamming distance} from programs in $P$ is bigger than $0$ (line 5). 
 
The Hamming distance between a term $t$ and the set of programs $P$ is calculated using 
the function $\func{hammDist}$, defined in \autoref{subfig:hammDist}.
The inputs to the function are a set of programs $P$ to compute the distance from, and a term $t$.
This function first 
determines the operator vector of $t$ (line 2) and the \emph{average} operator vector of all programs in $P$ (line 3). The final result is defined as the number of entries in the operator vector of $t$ whose value is greater than $\mathtt{OpTH}$, and the value of the corresponding entry in the average vector of $P$ is less than $\mathtt{OpTH}$.

For example, in \autoref{fig:motivate} (\#\textsc{iii}), the value assigned to the $\reg{alter}$ entry in the average operator vector of programs in $P$ is $0$, and hence, if $\mathtt{OpTH}$ is set to $1$, their Hamming distance from any term that has more than $1$ occurrences of $\reg{alter}$ is at least $1$; such terms will not be included in $v_3$.


\textbf{Semantic condensation}. The final step of pruning is to implement the beam-based cutoff of the state space based on the final ranking of programs with respect to the operator distribution. Unlike standard beam search methods, we do not perform a global ranking  on the search space when considering the beam. Instead, we maintain semantic variety in the beam exploration by ranking \emph{within} semantic equivalence classes of programs. Such \emph{condensing} of the set of programs within equivalence classes minimizes redundant syntactic variations of the same program in the search exploration which can come from a global ranking. 
At line 6, the prune function classifies terms in $v_3$ into \emph{semantic classes}. A semantic class is defined as a set of terms which have equal interpretations with respect to $\ex$. All terms in $v_3$ must belong to exactly one semantic class. The set of all semantic classes is stored in variable $v_4$ (line 6).

Next, using a call to function $\mathtt{ordEuc}$ (defined in \autoref{subfig:aux}), the procedure orders terms in each semantic class according to 
their syntactic similarity to the programs in $P$. The highest ranked programs in each semantic class are then identified and their union is stored in a variable $v_5$ (line 8). The number of terms selected from each class is determined by the size of that class and a tunable constant $\mathtt{BeamSize}$ (line 7). For example, assuming $\mathtt{BeamSize}$ is set to $2000$ and there are $5000$ terms in $v_3$, the top $400$ terms from a semantic class of size $1000$ will be selected to be in the pruned cache.

Finally, once the iteration over all sorts is finished, the procedure returns the fully pruned cache as the final result (line 10).

\subsection{Ranking the Synthesized Programs}
\label{subsec:rank}

The eventual goal of the CBS algorithm is to return a synthesized program to the user that satisfies the examples. But after a certain number of iterations of CBS in practice, there can be a large number of programs that satisfy the given examples. As in the other phases, our technique for ranking is also guided by considering similarity of the synthesized programs to the PTM candidates. The operator frequency  distribution as used above is a good signal for guiding the search in terms of which operator applications to explore, and is also a good indicator for the final preference of which program to pick from the set of synthesized programs. However, we also found that for final ranking it is helpful to use the additional stronger signal of direct string similarity of programs to the PTM candidates.

The function $\func{rank}$ is defined in \autoref{subfig:rank}. This function is called in the main function $\func{synthesize}$ (line 8). Given a set $T$ of terms  and a set $P$ of programs generated by the PTM, this function returns an ordered list of terms in $T$ according to their syntactic similarity to the programs in $P$. In particular, terms in $T$ are lexicographically ordered based on two different measures of distance from programs in $P$.
The first measure is the standard Euclidean distance between the operator vector of terms in $T$ and the average operator vector of programs in $P$ (line 2). This is to ensure that the final synthesized program is structurally as close as possible to the PTM's candidate programs. 

In order to distinguish terms in $T$ with the same Euclidean distance to $P$, the procedure next applies the Levenshtein distance~\cite{lev} as a more fine-grained measure of distance between (the string representation of) terms. The Levenshtein between two strings is defined as the minimum number of single character modifications required to transform one string into another. 

\subsection{Completeness}
\label{subsec:completeness}

We have described how our \textsc{nlx} algorithm implements a component-based synthesis  that is guided by the output of the PTM at every stage. While we empirically evaluate the effectiveness of these techniques in practice, in this section we consider what theoretical guarantees can be provided on completeness: can the algorithm eventually find a program if one exists? As is common for any program synthesis approach based on natural language input, completeness depends on the ability of the underlying language model, which in our case is the PTM, to find  relevant programs that match the intended task. Hence, we formulate a \emph{relative completeness} result with respect to the PTM output.  

Let $P$ be the candidate programs provided by the PTM. We define the \emph{closure}  of $P$ with respect to the DSL $\lang$, denoted $P_c$, as the set of all programs that can be constructed from all atomic components in $P$. Formally, $t \in P_c$ iff either $t$ is atomic (Definition \ref{def:subterm}) and $t \sqsubseteq p$ for some $p \in P$, or otherwise $t = op(t_1,\dots,t_n)$ where $op \in \op$ and $t_i \in P_c$. Hence, $P_c$ includes $P$ and all other programs that can possibly be constructed using components from $P$. We show that if the correct intended program exists in $P_c$, then our \textsc{nlx} algorithm can find a semantically equivalent program when given sufficient examples (under the assumption of a  condition of compositionality (T) holds for our DSL: for any terms $p$, $t$, $t'$, whenever 
$\denote{t}_E = \denote{t'}_E$, then
$\denote{p[t]}_E = \denote{p[t']}_E$).

\begin{corollary}[Relative completeness]
Let $P$ be the results of the PTM for a given natural language description $N$, and assume that the intended ground-truth program $p$ exists in the closure $P_c$ of $P$. Assume we set algorithm configuration parameter settings  $\mathtt{PrOcc} = 0$, $\mathtt{PrRed = 1}$ and have unbounded $\mathtt{SynthDepth}$, $\mathtt{BeamSize}$ and $\mathtt{OpTH}$. If compositionality condition (T) holds, then there exists a sufficient set of examples $E$ for which the \textsc{nlx} algorithm will return a program $p'$ that is semantically equivalent $p$.  
\end{corollary}

This relative completeness result follows from the fact that our \textsc{nlx} algorithm reduces to an exhaustive enumerative search under the extreme parameter settings above. Under the occurrence and redundancy probability settings $\mathtt{PrOcc} = 0$ and  $\mathtt{PrRed = 1}$ the cache is initialized with all possible components in $P$, which includes all atomic components. With unbounded iteration depth, unrestricted beam size and no constraints of operator frequency, the expansion phase explores all possible operator applications in the closure $P_c$. Assuming $p$ requires $k$ synthesis iterations to construct, let $p_1,...,p_n$ be all programs synthesized in up to $k$ iterations. Let $E$ be a set of examples that distinguishes $p$ from each of $p_1,...,p_n$ where $p_i \neq p$ (such an example set must exist or else $p$ will be semantically equivalent to some $p_i$). Then given the example set $E$, the algorithm will return the desired program $p$ after $k$ iterations. One notable issue is presented by our optimization on line 4 in \autoref{subfig:prune}, where we eliminate any term that is equivalent to any of its sub-terms. In case $p$ contains a term $t$ that is eliminated in favor of some semantically equivalent  $t'$, then we will synthesize $p'$ that uses  $t'$ instead of $t$, which will be semantically equivalent to $p$ (by condition (T)). 

%% file: Figs/algorithm.tex
\algsetup{
    linenosize=\Small,
    linenodelimiter = {:}
}
\begin{figure}
\begin{minipage}{0.5\textwidth}
\begin{subfigure}[b]{\textwidth}
\begin{mdframed}[backgroundcolor=grey12,linecolor=grey4,innerleftmargin=0.2cm]
\footnotesize
 \begin{algorithmic}[1]
  {\customline{}
  \STATE $\func{synthesize}_{\lang,\model}$ $(\natl,\ex):=$
  \STATE  \ \ \ \ \texttt{let} $\lang=(\sort,\const,\op,\mathtt{s}^\circ,\sigin,\sigret)$
  \STATE  \ \ \ \ $P:= \mathtt{exec}(\model,\natl)$
  \STATE  \ \ \ \ $\cache := \func{initialize}(P)$   
  \STATE  \ \ \ \ \textbf{foreach} \,$i$\, \textbf{in} $\{1,2,\dots,\mathtt{SynthDepth}\}$:
  \STATE  \ \ \ \ \ \ \ \ \ $\cache:=\func{expand}(\cache,P,\ex)$
  \STATE  \ \ \ \ $v_1 := \{t \ALT t\!\in\! \cache(\mathtt{s}^\circ) \wedge t\models \ex\}$
  \STATE  \ \ \ \ \textbf{return} $\func{rank}(v_1,P).\mathtt{top}(1)$
  }
 \end{algorithmic}
 \end{mdframed}
 \vspace{-1.2mm}
   \caption{main function}
    \vspace{0.7mm}
 \label{subfig:synthesize}
 \end{subfigure}
\\
\begin{subfigure}[b]{\textwidth}
\begin{mdframed}[backgroundcolor=grey12,linecolor=grey4,innerleftmargin=0.2cm]
\footnotesize
 \begin{algorithmic}[1]
 {\customline{}
  \STATE $\func{initialize}$ $(P):=$
  \STATE  \ \ \ \ $v_1 := \{t \ALT t\!\sqsubseteq\! p \wedge p\!\in\!P\}$
  \STATE  \ \ \ \ $v_2 := \{t \ALT t\in v_1\wedge \mathtt{cnt}(t,P) / |P| \geq \mathtt{PrOcc}\}$
  
  \STATE  \ \ \ \ $v_3 := \emptyset$ 
 \STATE  \ \ \ \ \textbf{foreach} $t$ \textbf{in} $v_2$  
  \STATE  \ \ \ \ \ \ \ \  $s\, :=  \{t' \ALT t'\in v_1 \wedge t\!\sqsubseteq\! t'\}$
 \STATE   \ \ \ \ \ \ \ \  $s_r :=  \{t' \in s \ALT  t \!\neq\! t' \wedge \mathtt{cnt}(t,P)\!=\!\mathtt{cnt}(t'\!,P) \}$
\STATE  \ \ \ \ \ \ \ \ \textbf{if} $|s_r|/|s| \leq \mathtt{PrRed}$ \textbf{ then}
  \STATE  \ \ \ \ \ \ \ \ \ \ \ \  $v_3\, := v_3 \cup \{t\}$
  
  \STATE  \ \ \   $v_3\, := v_3 \cup \mathtt{stdComps(\!\lang)}$
 
 \STATE  \ \ \ \textbf{foreach} $s$ \textbf{in} $\sort$:
  \STATE   \ \ \ \ \ \ \ $\cache(s) := \{t\!\ALT\! t\!\in\! v_3 \wedge \mathtt{srt}(t)\!=\!s\}$
  \STATE   \ \ \ \textbf{return} $\cache$
  }
 \end{algorithmic}
 \end{mdframed}
  \vspace{-1.2mm}
\caption{cache initialization}
 \vspace{0.5mm}
  \label{subfig:init}
 \end{subfigure}
 \\
\begin{subfigure}[b]{\textwidth}
\begin{mdframed}[backgroundcolor=grey12,linecolor=grey4,innerleftmargin=0.2cm]
\footnotesize
 \begin{algorithmic}[1]
  {\customline{}
  \STATE $\func{expand}$ $(\cache, P, \ex):=$
 \STATE \ \ \ \ $\cache':=\func{prune}(\cache,P, \ex)$
  \STATE  \ \ \ \ \textbf{foreach} $op$ \textbf{in} $\op:$  
  \STATE  \ \ \ \ \ \ \ \  $<\!\!s_1,\dots,s_n\!\!>\, := \sigin(op)$
  \STATE  \ \ \ \ \ \ \ \  $v_1 := \{op(t_1,\dots,t_n) \ALT \forall_{1\leq i\leq n}\;t_i\!\in \cache'(s_i)\}$
    \STATE  \ \ \ \ \ \ \ \  $s := \sigret(op)$ 
  \STATE  \ \ \ \ \ \ \ \ $\cache := \cache[s\mapsto \cache(s)\!\cup\! v_1]$
  \STATE  \ \ \ \ \textbf{return} $\cache$
  }
 \end{algorithmic}
 \end{mdframed}
  \vspace{-1.2mm}
 \caption{cache expansion}
  \vspace{0.5mm}
  \label{subfig:expand}
 \end{subfigure}
\\
\begin{subfigure}[b]{\textwidth}
\begin{mdframed}[backgroundcolor=grey12,linecolor=grey4,innerleftmargin=0.2cm]
\footnotesize
 \begin{algorithmic}[1]
  {\customline{}
  \STATE $\func{prune}$ $(\cache, P, \ex):=$ 
  \STATE \ \ \ \ \textbf{foreach} $s$ \textbf{in} $\sort$:
  \STATE \ \ \ \ \ \ \ \ $v_1 := \cache(s)$ 
  
  \STATE  \ \ \ \ \ \ \ \  $v_2 := \{t \ALT t\!\in\! v_1 \wedge \forall_{t'\sqsubseteq t}\; \denote{t}_\ex\not\eq\denote{t'}_\ex  \}$
  \STATE  \ \ \ \ \ \ \ \  $v_3 := \{t \ALT t\!\in\! v_2 \wedge \func{hammDist}(P,t) \!=\! 0 \}$
  
  \STATE \ \ \ \ \ \ \ \ $v_4 := \{T \!\ALT\!  T\!\subseteq\! v_3 \wedge (\forall_{t,t'\in T}.\, \denote{t}_\ex \!=\! \denote{t'}_\ex) \wedge $
  
  \hspace{27.3mm} $(\forall_{t\in v_3} \exists_{T'\in v_4}.\; t\!\in\!T')\}$
  
  \STATE \ \ \ \ \ \ \ \ $f:= \lambda T.\; (|T|/|v_3|\times\mathtt{BeamSize})$

  \STATE \ \ \ \ \ \ \ \ $v_5 := \bigcup_{T\in v_4}\{t \!\ALT\! t\!\in\! \mathtt{ordEuc}(T,P).\mathtt{top}(f(T))    \}$ 
  \STATE \ \ \ \ \ \ \ \ $\cache := \cache[s\mapsto v_5]$
  \STATE \ \ \textbf{return} $\cache$
  }
  \end{algorithmic} 
 \end{mdframed}
  \vspace{-1.2mm}
 \caption{cache pruning}
 \label{subfig:prune}
 \end{subfigure}
\end{minipage}
%
%
%
\hfill
%
%
%
%
\begin{minipage}{0.49\textwidth}
\begin{subfigure}[b]{\textwidth}
\begin{mdframed}[backgroundcolor=grey12,linecolor=grey4,innerleftmargin=0.2cm]
\footnotesize
 \begin{algorithmic}[1]
  {\customline{}
  \STATE $\func{rank}$ $(T,P):=$ 
  \STATE  \ \ \ \ $f_1 = \lambda t.\; \func{eucDist}(P,t)$ 
  \STATE  \ \ \ \ $f_2 = \lambda t.\; (\sum_{p\in P}\mathtt{lev}(t,m))/|P|)$ 
  \STATE  \ \ \ \ \textbf{return} $T.\mathtt{orderBy}(f_1,f_2)$
  }
 \end{algorithmic}
 \end{mdframed}
  \vspace{-1.2mm}
 \caption{ranking candidates}
  \vspace{3mm}
  \label{subfig:rank}
 \end{subfigure}
\\
\begin{subfigure}[b]{\textwidth}
\begin{mdframed}[backgroundcolor=grey12,linecolor=grey4,innerleftmargin=0.2cm]
\footnotesize
 \begin{algorithmic}[1]
  {\customline{}
  \STATE $\func{eucDist}$ $(P,t):=$ 
  \STATE  \ \ \ \ $<\!\!v_1,\dots,v_n\!\!> := \mathtt{opVec}(t)$
  \STATE  \ \ \ \ $<\!\!v'_1,\dots,v'_n\!\!> := (\sum_{p\in P} \mathtt{opVec}(p))/|P|$
  \STATE  \ \ \ \ \textbf{return} $\sqrt{\sum_{1\leq i\leq n}(v_i-v'_i)^2}$
  }
 \end{algorithmic}
 \end{mdframed}
  \vspace{-1.2mm}
 \caption{standard Euclidean distance}
  \vspace{2mm}
  \label{subfig:eucDist}
 \end{subfigure}
\\
\begin{subfigure}[b]{\textwidth}
\begin{mdframed}[backgroundcolor=grey12,linecolor=grey4,innerleftmargin=0.2cm]
\footnotesize
 \begin{algorithmic}[1]
  {\customline{}
  \STATE $\func{hammDist}$ $(P,t):=$ 
  \STATE  \ \ \ \ $<\!\!v_1,\dots,v_n\!\!> := \mathtt{opVec}(t)$
  \STATE  \ \ \ \ $<\!\!v'_1,\dots,v'_n\!\!> := (\sum_{p\in P} \mathtt{opVec}(p))/|P|$
  
  \STATE  \ \ \ \ \textbf{return} $|\{i \ALT v_i > \mathtt{OpTH} \wedge  v'_i < \mathtt{OpTH} \wedge\; 1\!\leq \!i\!\leq\! n\}|$
  }
 \end{algorithmic}
 \end{mdframed}
  \vspace{-1.2mm}
 \caption{customized Hamming distance}
 \vspace{3mm}
  \label{subfig:hammDist}
 \end{subfigure}
\\
\begin{subfigure}[b]{\textwidth}
\scriptsize
\begin{mdframed}[backgroundcolor=grey12,linecolor=grey4,innerleftmargin=0.2cm]
{\renewcommand{\arraystretch}{1}%
\begin{tabular}{p{1.5cm}p{4.2cm}}
  $\mathtt{SynthDepth}:$ & \scriptsize number of synthesis iterations  \\    \arrayrulecolor{grey8}\hline
 \vspace{  0.06mm}$\mathtt{PrOcc}:$  &\scriptsize   {minimum probability of occurrence of a component in programs in $P$}  \\    \arrayrulecolor{grey8}\hline
\vspace{  0.06mm}$\mathtt{PrRed}:$  &\scriptsize   {maximum probability of redundancy of a component in programs in $P$}  \\    \arrayrulecolor{grey8}\hline
  \vspace{  0.06mm} $\mathtt{BeamSize}:$ &\scriptsize   { number of programs (of each sort) used to synthesize the next set} \\     \arrayrulecolor{grey8}\hline
   \vspace{0mm} $\mathtt{OpTH}:$  & \scriptsize    { threshold that determines low-frequency operator occurrence} 
\end{tabular}
}
\end{mdframed}
\caption{constants}
\label{subfig:aux}
 \end{subfigure}
\\[3mm]
\begin{subfigure}[b]{\textwidth}
\scriptsize
\begin{mdframed}[backgroundcolor=grey12,linecolor=grey4,innerleftmargin=0.2cm]
{\renewcommand{\arraystretch}{0.9}%
\begin{tabular}{p{1.5cm}p{4.2cm}}
  \vspace{  0mm}$\mathtt{opVec}(t):$ &\scriptsize  { returns a vector composed of the number of occurrences of each DSL operator in $t$} \\     \arrayrulecolor{grey8}\hline
   \vspace{  0.06mm}$\mathtt{orderBy}(f_1,f_2)$:&\scriptsize   { returns a lexicographically ordered list based on given score functions $f_1$ and $f_2$}\\     \arrayrulecolor{grey8}\hline
  \vspace{  0.06mm} $\mathtt{lev}(t_1,t_2)$:& \scriptsize   {returns the Levenshtein distance between string representations of $t_1$ and $t_2$ } \\     \arrayrulecolor{grey8}\hline
 \vspace{  0.06mm} $\mathtt{top}(n)$:&\scriptsize  { returns the first $n$ elements from an ordered list}  \\     \arrayrulecolor{grey8}\hline
 $\mathtt{srt}(t)$:& \scriptsize   {returns the sort of the root operator in $t$ } \\     \arrayrulecolor{grey8}\hline \vspace{0.06mm}
 $\mathtt{cnt}(t,P)$:&\scriptsize  {returns number of programs $p$ in $P$ such that $t \sqsubseteq p$}
 \\     \arrayrulecolor{grey8}\hline
 \vspace{  0.01mm} $\mathtt{exec}(\model,\natl)$:&\scriptsize  { Runs $\model$ on $\natl$ and returns the resulting candidate programs}
 \\      \arrayrulecolor{grey8}\hline
 \vspace{  0.06mm}
 $\mathtt{stdComps(\!\lang)}$:&\scriptsize  {the standard components to include for DSL $\lang$}
  \\      \arrayrulecolor{grey8}\hline
 \vspace{  0.06mm}
 $\mathtt{ordEuc}(T,P)$:&\scriptsize  {orders terms in $T$ based on their Euclidean distance from average program in $P$}
\end{tabular}
}
\end{mdframed}
\caption{auxiliary functions}
\label{subfig:aux}
 \end{subfigure}
%
%
%
%
\end{minipage}
%
%
%
%
\vspace{-1mm}
  \caption{\textsc{nlx} algorithm for multi-modal program synthesis, parameterized on a DSL $\lang$ and PTM $\model$  
  }
  \label{fig:alg}

\end{figure}

%% file: opt_new.tex
\section{Optimized use of the PTM}
\label{sec:opt}

\ignore{
Structure:
\begin{itemize}
  \item What is the problem? Motivate the formative study.
  \item What are the parameters we are studying and methodology?
  \item Results
\end{itemize}
}

Section~\ref{sec:synth} describes a generic component-based synthesis
technique that uses the top candidate results of a PTM to seed and guide
an enumerative search.
The effectiveness of this process, however, depends on the quality of the
initial results received from the PTM.
%
%
Getting the right results from the PTM is heavily dependent on \emph{asking
the questions in the right way.}
In particular,
using the PTM effectively involves three distinct steps.
First, the task at hand is encoded into a \emph{prompt} that acts as the
context for the PTM.
%
Then, the prompt is provided as input to the PTM which then produces a
\emph{completion}.
Finally, the candidate program from the output completion is \textit{extracted}.
In this section, we explain each of these steps in details and conduct a formative study to design a technique to
generate high quality prompts to obtain useful initial programs.

\input{Figs/prompt_fig}

Following~\cite{brown2020language}, we use the PTM as a \textit{few-shot learner},
i.e. the model is provided a few question-answer pairs that act as examples
of the task at hand.
Note that we use the term question-answer pair (instead of example) to avoid confusion
with the examples required for the synthesis tasks given to \textsc{nlx} algorithm.
For instance,
\autoref{subfig:prompt} shows an example prompt outlining the prompt structure that is used as context for the PTM.
The prompt consists of three parts:
\begin{enumerate*}[label=(\roman*)]
  \item a high level description of the task domain (lines~1-3),
  \item a sequence of sample question-answer pairs (lines~5-15), and
  \item the question of interest (line~17).
\end{enumerate*}
%
%

%
Structuring the prompt in this way has multiple advantages.
First, the question-answer pairs often contain components that increase the
probability of the PTM returning results using those components, e.g. \texttt{vowel} is used in a question-answer pair (line~5) which is also part of the final question. 
Additionally, as a small advantage, the structured prompt biases the PTM to
produce a response in the same format making the task of extracting the
resulting program as simple as picking the right \emph{stop sequence} (here,
\texttt{NL:}).

For example, \autoref{subfig:completion} presents \REV{one of the completions}\footnote{
    Note that GPT-3 is non-deterministic by nature.
    Completions shown represent typical results for the prompts.
  } that GPT-3 produces given the prompt in \autoref{subfig:prompt}. The completion consists of a candidate program for the task (line~18) and 
  a few additional lines, following the same pattern from the prompt (lines~19-21). It is easy to see how the candidate program can be extracted from the completion using the stop sequences. 
    Although in this example the PTM was able to successfully generate the intended program, that is not always the case. For example, if we remove the first two question-answer pairs from the prompt (lines~5-10), the completion produced by GPT-3 does not solve the task correctly and returns  programs like \texttt{[A-Z]\{3\}[0-9]\{4\}.*}, which does not even include the correct components of the intended program. 
  
  The above example highlights the main challenges when using PTMs as program synthesizers. In the remaining of this section, we will present our prompt generation approach to maximize the likelihood of getting the correct programs with right components in the PTM's completion.

  %
  %
  %
  %
  %
  %

\subsection{Formative Study on Selecting Question-Answer Pairs}

Most PTMs restrict the size of the input prompt they accept.
%
For example, the GPT-3 prompt is restricted to 2048 tokens (i.e. small units that are meaningful and occur more generally).
Given this limited prompt size, choosing the right set of question-answer
pairs to act as examples for $k$-shot learning becomes very important for the
quality of results.
We introduce a technique for choosing relevant question-answer pairs, and
study multiple variations to determine the optimal parameters for prompt
generation.
\begin{algorithm}[t]
  \SetAlCapFnt{\scriptsize}
  \SetAlCapNameFnt{\scriptsize}
  \footnotesize
  \begin{algorithmic}[1]
    \REQUIRE Question-answer corpus $\mathsf{QA} = (q_0, a_0), \ldots , (q_n, a_n)$
    \REQUIRE Natural language description at hand $q^*$
    \REQUIRE Relevance metric $\mathcal{R}$
    \REQUIRE Result size threshold $k \in \mathbb{R}$ and similarity threshold $t \in \mathbb{R}$
    \STATE $\mathsf{RelevantQA} \gets \text{empty sequence}$
    \WHILE {$\mathsf{QA} \neq \emptyset \land |\mathsf{RelevantQA}| < k$}
    \STATE $(q_m, a_m) \gets \mathsf{argmax}_{(q_i, a_i) \in \mathsf{QA}} \mathcal{R}(q^*, q_i)$\label{line:max}
    \STATE $\mathsf{QA} \gets \mathsf{QA} \setminus \{ (q_m, a_m) \}$
    \IF {$\not \exists (q, a) \in \mathsf{RelevantQA}.\ \mathsf{LevenshteinDistance}(a, a_m) \!<\! t\;$} \label{line:similarity}
    \STATE $\mathsf{RelevantQA} \gets \mathsf{RelevantQA}; (q_m, a_m)$
    \ENDIF
    \ENDWHILE
    \RETURN $\mathsf{RelevantQA}$
  \end{algorithmic}
  \caption{Question-Answer pair ranking}
  \label{algo:prompt-selection}
\end{algorithm}

Algorithm~\ref{algo:prompt-selection} depicts our question-answer pair
selection technique.
The primary inputs are
\begin{enumerate*}
  \item a corpus of question answer pairs, $\mathsf{QA} = (q_0, a_0), (q_1,
    a_1), \ldots (q_n, a_n)$, where each question $q_i$ is a natural language
    description and the answer $a_i$ is the corresponding program,
    and
  \item a question $q^*$ that represents the task in hand.
\end{enumerate*}
The procedure  returns a sequence $\mathsf{RelevantQA} =
(q_{i_0}, a_{i_0}), \ldots, (q_{i_k}, a_{i_k})$ of $k$ question-answer pairs to be used in the prompt.
The algorithm is parameterized by a relevance metric $\mathcal{R}$ on questions. A greater $\mathcal{R}(q,q')$ score indicates that question $q$ (and its answer) is more relevant to (answering) $q'$. 
At a high level, Algorithm~\ref{algo:prompt-selection} orders the available question-answer pairs in $\mathsf{QA}$ based on their relevance to $q^*$ and identifies the highest ranked question-answer pair (line~\autoref{line:max}). The chosen pair $(q_m, a_m)$ is added to the result sequence if $a_m$ is
not ``too close'' to an already selected answer in $\mathsf{RelevantQA}$ (line~\ref{line:similarity}).
Here we define closeness of answers as the Levenshtein distance between them, which is a fine-grained measure of distance between strings at the level of characters~\cite{lev}. If the distance is less than a threshold $t$ then the answers are considered too close and the question-answer pair is discarded.
This is to ensure that the PTM is not biased toward a particular group of tasks and does not produce sub-optimal results. 

Below, we introduce two  classical metrics of relevance from 
the information retrieval literature and study the impact of each  on the quality of the PTM's completion.

\subsubsection{Relevance Metrics}
Suppose we are interested in computing $\mathcal{R}(q, q')$, the relevance  of question $q$ to question $q'$.
%
We use $|q|$ and $|q'|$ to denote the number of tokens in $q$ and $q'$,
respectively, and define $\mathsf{CT}(q,q')$ to be the set of tokens common to $q$ and
$q'$. We now introduce two different definitions for $\mathcal{R}(q,q')$:

\begin{enumerate}[topsep=0pt,itemsep=-1ex,partopsep=1ex,parsep=1ex]
  \item[-] {\bf Token match:}
    This metric, $\mathcal{R}_{\mathsf{TM}}$, measures the fraction of the number of tokens in $q$
    that are also present in $q'$, i.e.  $\mathcal{R}_{\mathsf{TM}(q, q')} = \frac{| \mathsf{CT}(q,q') |}{|q|}$.
  \item[-] {\bf TF-IDF:} The measure 
  $\mathcal{R}_{\mathsf{TM}}$ treats all tokens identically because we just count the tokens. 
  However, rare tokens are better indicators of relevance. We follow the standard term-frequency inverse document frequency (TF-IDF)
    technique~\cite{sparckjones1972} to increase weight of rare tokens.
    In particular, we define the TF-IDF score of each token and weight them based on this score.
    The score $\mathsf{TFIDF}(T)$ of a token $T$ is the product of
    \begin{enumerate*}
      \item the \emph{term frequency} of $T$, i.e., the number of times $T$
        occurs in $q$, and
      \item the log of the \emph{inverse document frequency} of $T$, i.e.,
        the negative log of the fraction of questions from the corpus that $T$
        appears in.
    \end{enumerate*}
    Thus, we have $\mathcal{R}_{\mathsf{TFIDF}}(q,q') =
    \frac{\sum_{T\in CT(q,q')} \mathsf{TFIDF}(T)}{ \sum_{T\in q} \mathsf{TFIDF}(T)}$.
\end{enumerate}



\subsubsection{Experimental Results}
In this part, we study the impact of the similarity check in
Algorithm~\ref{algo:prompt-selection} (line~\ref{line:similarity}) on the recall of the PTM for different relevance metrics.
We use GPT-3 as our PTM with the \textit{temperature parameter} set {\color{black} to
$0.6$. In GPT-3 terminology, temperature 0.0 represents an entirely deterministic value, whereas 1.0 represents output that is fully stochastic. In the domain of regular expression, we aimed to allow the PTM sufficient randomness to generate a varied candidate set, but not an entirely random one such that the similar components between candidates indicated some measure of confidence. After some brief initial trials, 0.6 was selected for temperature} and $10$ as the threshold on the
number of question-answer pairs in the prompt.
\begin{wraptable}[12]{r}{0.33\textwidth}
\center
  \footnotesize
  \begin{tabular}{c c r}
    \hline
    \begin{tabular}{@{}c@{}}Relevance \\ Metric\end{tabular}
       & \begin{tabular}{@{}c@{}}Similarity \\ Check\end{tabular} &   \begin{tabular}{@{}c@{}}Top-20 \\ Recall\end{tabular}   \\
    \hline
    \REV{Hand-Picked}  & \REV{No}              &             \REV{0.32} \\
    Random  &  No              &             0.33 \\
    $\mathcal{R}_{\mathsf{TFIDF}}$            &  Yes             &             0.46 \\
    $\mathcal{R}_{\mathsf{TFIDF}}$           &  No              &             0.44 \\
    $\mathcal{R}_{\mathsf{TM}}$       &  Yes             &             \REV{0.42} \\
    $\mathcal{R}_{\mathsf{TM}}$       &  No              &             \REV{0.40} \\
  \end{tabular}
  \vspace{2mm}
\caption{Recall within top 20 completions for variants of Algorithm~\ref{algo:prompt-selection}.}
\label{tab:prompt-selection-variants}
\end{wraptable}
%
Our corpus contained 4855 question-answer pairs from the \cite{deepregex} dataset (see
Section~\ref{sec:eval}) and our test tasks consisted of
115 questions from the same dataset.
For each variant of Algorithm~\ref{algo:prompt-selection}, we generated a
prompt based on the relevant pairs returned by the variant and measured the
recall in top 20 completions, i.e., in what fraction of the cases is the
correct answer is in the top 20 completions produced by
the PTM.
For the baseline, \REV{we propose two techniques: First, a straw-man procedure that randomly selects
question-answer pairs from the corpus for each question, Second a Hand-Picked context which remains unchanged throughout the entire experiment}. 
%
%
    First, we fix the threshold $k$ on the size of the result to be
    10, and test variants using token match and TF-IDF metrics
    and with and without the Levenshtein distance based similarity check.
    %
%
The results are summarized in Table~\ref{tab:prompt-selection-variants}.
%
The results highlight two key points:
\begin{enumerate*}
  \item Intelligent relevance-based selection of question-answer pairs for
    the in-context $k$-shot learning makes a significant difference to the
    recall of the PTM.
  \item Using the Levenshtein distance based similarity check increases the
    diversity in the question-answer pairs used in the prompt, and thereby
    increases the recall of the PTM.
\end{enumerate*}

Based on the  above insights, we chose the best variation with TF-IDF relevance metrics and
the similarity check for conducting experiments in Section~\ref{sec:eval}.

%% file: Figs/prompt_fig.tex
\begin{wrapfigure}{r}{0.5\textwidth}
\center
\begin{subfigure}{0.47\textwidth}
\begin{mdframed}[linecolor=grey4,innerleftmargin=0.1cm]
  \scriptsize
  \begin{alltt}
{\color{grey5}(01)} Here are some examples of regular expressions 
{\color{grey5}(02)} and their descriptions. Use them to generate a
{\color{grey5}(03)} regular expression that matches the description.
{\color{grey5}(04)}
{\color{grey5}(05)} NL: lines which begin with an upper case vowel
{\color{grey5}(06)} Regex: [AEIOU].*
{\color{grey5}(07)} 
{\color{grey5}(08)} NL: match lines which contain only consonants
{\color{grey5}(09)} Regex: [^AEIOUaeiou]*
{\color{grey5}(10)} 
{\color{grey5}(11)} NL: lines ending with a digit followed by period
{\color{grey5}(12)} Regex: .*[0-9][.]
{\color{grey5}(13)} 
{\color{grey5}(14)} NL: dates in ISO 8601 format
{\color{grey5}(15)} Regex: [0-9]\{4\}-[0-9]\{2\}-[0-9]\{2\}
{\color{grey5}(16)} 
{\color{grey5}(17)} NL:  lines starting with three upper case vowels 
         followed by four digits
{\color{grey5}(18)} Regex:
  \end{alltt}
  \vspace{-2mm}
  \end{mdframed}
  \caption{prompt}
  \label{subfig:prompt}
  \end{subfigure}
 %
 %
 %
 %
 \\[1mm]
  \begin{subfigure}{0.47\textwidth}
  \begin{mdframed}[linecolor=grey4,innerleftmargin=0.1cm]
  {\scriptsize
  \begin{alltt}
{\color{grey5}(18)} Regex: \textcolor{purple}{[AEIOU]\{3\}[0-9]\{4\}.*}
{\color{grey5}(19)}
{\color{grey5}(20)} \textcolor{purple}{ NL: lines starting with a digit followed by three}
         \textcolor{purple}{ upper case letters followed by two digits}
{\color{grey5}(21)} \textcolor{purple}{Regex: [0-9][A-Z][A-Z][A-Z][0-9]\{2\}}
  \end{alltt}
}
\vspace{-2mm}
 \end{mdframed}
  \caption{completion}
    \label{subfig:completion}
  \end{subfigure}
 \vspace{-1mm}
\caption{A prompt and the corresponding completion.
Line numbers in parenthesis are for illustration only}
\label{fig:prompt}
\end{wrapfigure}

%% file: eval.tex
\section{evaluation}
\label{sec:eval}

This section presents an  empirical evaluation of our synthesis approach across two programming domains. 
First, in §\ref{subsec:regex} we present \textsc{nlx-reg} -- an implementation of our algorithm for the domain of regular expressions, and compare it to 
the state-of-the-art regular expression synthesizer.
Next, in §\ref{subsec:css},
we introduce \textsc{nlx-css} for the domain of CSS selectors and evaluate it on 
a corpus of standard synthesis tasks in this domain.


\subsection{Domain of Regular Expressions}
\label{subsec:regex}


We now present our synthesizer for regular expressions from natural language and examples. This synthesizer, named  \textsc{nlx-reg}, implements an instance of the domain-agnostic algorithm presented in \autoref{sec:synth}, and is written in C\# language with about $5k$ lines of code. 
We apply 
\textsc{nlx-reg} on a set of synthesis benchmarks adopted from various sources and assess its performance by comparing it to 
three baselines, including the state-of-the-art synthesizer for regular expressions.
We begin by describing these  baseline systems: 

\begin{enumerate}
    
 \item \textbf{REGEL} is a tool developed by \citet{regel}, and is the state-of-the-art synthesizer for regular expressions
    from natural language and examples. 
    REGEL works by first generating a sketch (i.e. a basic scaffolding) of the target expression from the given English description of the task, and then
    completes the sketch using an enumerative search guided by the given examples. REGEL is designed specifically for the domain of regular expressions and cannot be applied to other DSLs. 
    \citet{regel} report a significantly higher accuracy rate for REGEL ($80\%$ vs $43\%$) compared to DeepRegex~\cite{deepregex}, the prior state-of-the-art tool for generating regular expressions directly from natural language.
    The DSL of regular expressions used in DeepRegex and REGEL is similar to
    ours except that, for implementation reasons, our DSL does not include the $\mathtt{And}$ (intersection) and $\mathtt{Not}$ (complement) operators, which are not supported by many standard libraries.

    \item \textbf{GPT-3} represents the next baseline system in our setup, which is simply a PTM that is used as an end-to-end synthesis tool.
    In other words, the top candidate generated for each task by the PTM is compared to the ground-truth without any further processing.

 \item \textbf{BFS} represents the \emph{brute force search} approach of component-based synthesis: it implements an exhaustive bottom-up search that starts with the initial set of \textit{all} atomic components found in \textit{any} of the PTM candidates, and applies all DSL operators at every iteration of the search. This baseline represents a simple way of combining the PTM output with component based synthesis, as opposed to the techniques for initialization, expansion and ranking that we introduced in \autoref{sec:synth} and are implemented in our \textsc{nlx-reg} system.  
    
\end{enumerate}


\begin{figure}[t]
\footnotesize
\hfill
\begin{subfigure}[b]{0.69\textwidth}
\begin{mdframed}[backgroundcolor=grey12,linecolor=grey4,innerleftmargin=0.3cm]
\textbf{Question:}
I want to validate decimal values with up to 18 digits before the decimal and 1 digit after; with the decimal point and the digit after it being optional.
For example all the following three numbers should be accepted:
$\mathtt{100.1}$,
$\mathtt{123456789.2}$, and
$\mathtt{123456789}$.
But these three numbers should not:
$\mathtt{1.01}$,
$\mathtt{1234567891234567891}$,  and
$\mathtt{1234567891234567891.0}$. 

I am currently using 
$(\![0\!-\!9]\!\{1,18\})\!+(\backslash.[0\!-\!9]\!\{1\})?$
as my regular expression, however it seems to be accepting things that are more than 18 digits before the decimal point.
Does anyone know what I did wrong here?

\textbf{Answer:}

Drop the '+':
$(\![0\!-\!9]\!\{1,18\})(\backslash.[0\!-\!9]\!\{1\})?$
\end{mdframed}
\caption{}
\label{subfig:post}
\end{subfigure}
\hfill
\begin{subfigure}[b]{0.3\textwidth}
\begin{mdframed}[backgroundcolor=grey12,linecolor=grey4,innerleftmargin=0.3cm]
\footnotesize
\textbf{Natural Language}: 
    
   A regex which accepts numbers up to 18 digits and an optional decimal point followed by a digit at the end

\textbf{Examples}: 
    \\
    $(\mathtt{100.1},\top)$,  \;$(\mathtt{123456789.2},\top)$,
    \\
    $(\mathtt{123456789},\top)$, \;$(\mathtt{1.01},\bot)$,\\
    $(\mathtt{1234567891234567891},\bot)$,
    \\
    \quad $(\mathtt{1234567891234567891.0},\bot)$
\end{mdframed}
\caption{}
\label{subfig:task}
\end{subfigure}
\hfill
\vspace{-2mm}
\caption{A StackOverflow post (left) and the extracted task (right)}
\label{fig:so}
\end{figure}

We applied  
\textsc{nlx-reg} and all the above baseline systems
on two sets of synthesis tasks and we will report the accuracy of each system on both sets and also per each set separately. Following is a summary of how we curated each of these benchmark sets:

    \subsubsection*{\textbf{DeepRegex}} 
        \citet{regel} originally evaluated REGEL using a set of 200 synthesis tasks sampled from DeepRegex benchmark set~\cite{deepregex}. 
        DeepRegex consists of 10000 pairs of natural language descriptions and regular expressions, automatically generated 
        using a small manually-crafted grammar.
        The artificially created natural language descriptions are then paraphrased through crowd-sourcing.         
        Since tasks in DeepRegex set only include English descriptions, Chen et al. also asked users to provide examples (4 positive and 5 negative on average) for each task.
        We eliminated 75 tasks where the ground-truth required either $\mathtt{And}$ or $\mathtt{Not}$ operators, which are not supported by \textsc{nlx-reg} as mentioned above. 
        We used the remaining tasks for evaluating all systems. 


\subsubsection*{\textbf{StackOverflow}}      
To complement the DeepRegex benchmarks with more challenging real-world scenarios, 
we also curated a set of synthesis tasks based on questions submitted to StackOverflow online forum. We initially retrieved posts tagged with keywords "regex" and "regular expression" and  identified  
    cases with exactly one regular expression in the question, $r_q$, and one regular expression in the accepted answer, $r_a$. 
    Using these expressions, we were able to  \textit{automatically} generate positive and negative examples for each task. In particular, we executed both expressions on the body of the question and collected all strings accepted by $r_a$ (i.e. the ground-truth) as positive examples. Similarly, all strings
    accepted by $r_q$ and
    rejected by $r_a$ were collected as negative examples.

    As a concrete example, consider \autoref{subfig:post} which presents a StackOverflow post\footnote{\href{https://stackoverflow.com/questions/19746891}{https://stackoverflow.com/questions/19746891}} identified using the above procedure. 
      The question in this post explains a task using a combination of natural language and positive and negative examples. It also 
      provides a faulty expression (i.e. $r_q$) which does not correctly perform that task. The accepted answer includes the correct expression for the task (i.e. $r_a$).
      Note that all positive examples provided by the user are accepted by $r_a$ and all negative examples are accepted by $r_q$ and rejected by $r_a$.
        The task extracted from this post is shown in \autoref{subfig:task}. 
      While we were able to automatically extract examples for each task, we relied on users across our institution to read the posts and paraphrase them concisely to eliminate redundancies common in online posts. 
    Using this  methodology, we collected a set of 25 tasks with an average of 4.3 positive  and 1.4 negative examples per task.

\begin{figure}[t]
\hfill
\begin{subfigure}{0.32\textwidth}
\includegraphics[width=\textwidth]{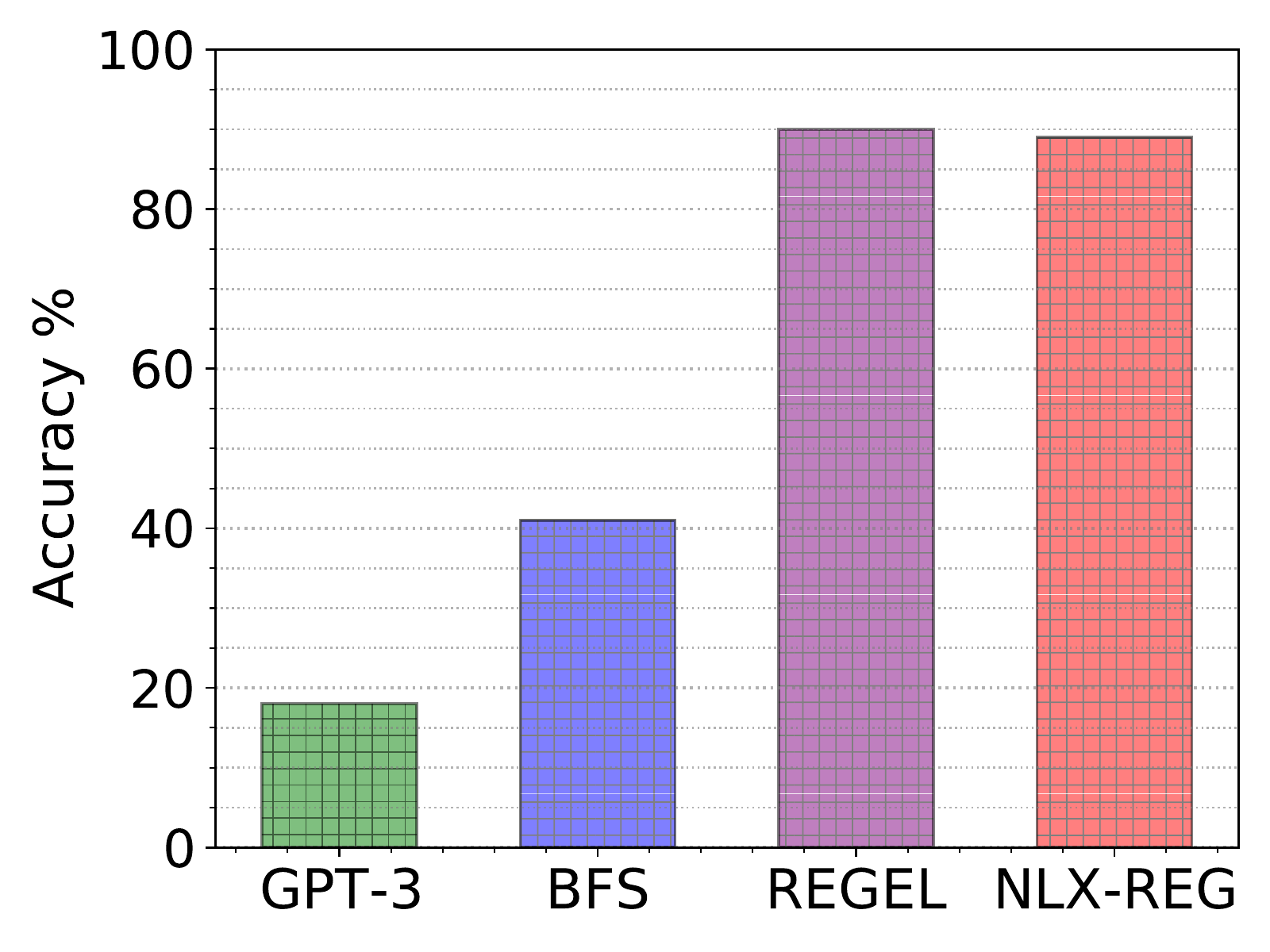}
\caption{DeepRegex}
\label{subfig:dr}
\end{subfigure}
\hfill
\begin{subfigure}{0.32\textwidth}
\includegraphics[width=\textwidth]{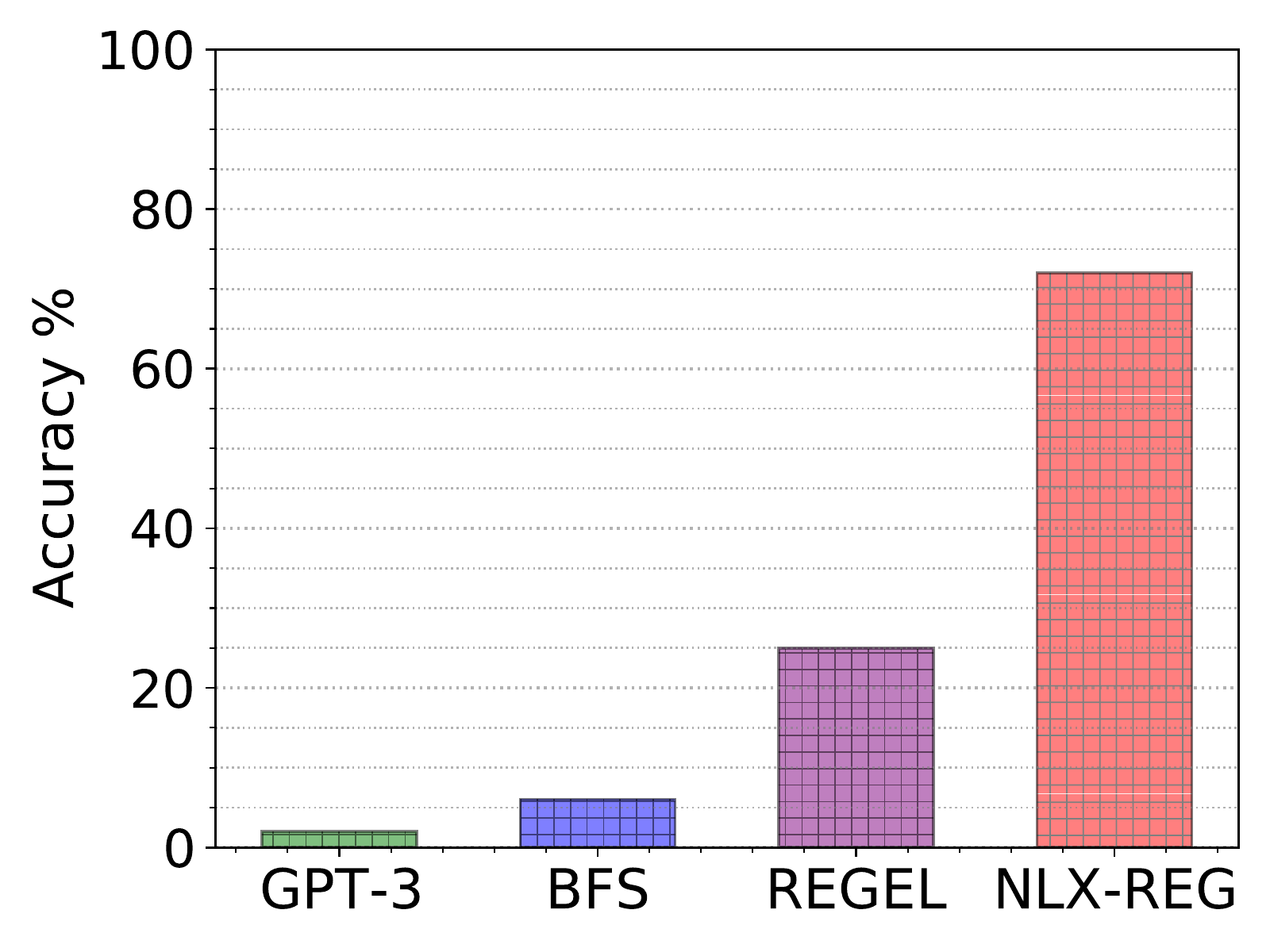}
\caption{StackOverflow}
\label{subfig:so}
\end{subfigure}
\hfill
\begin{subfigure}{0.32\textwidth}
\includegraphics[width=\textwidth]{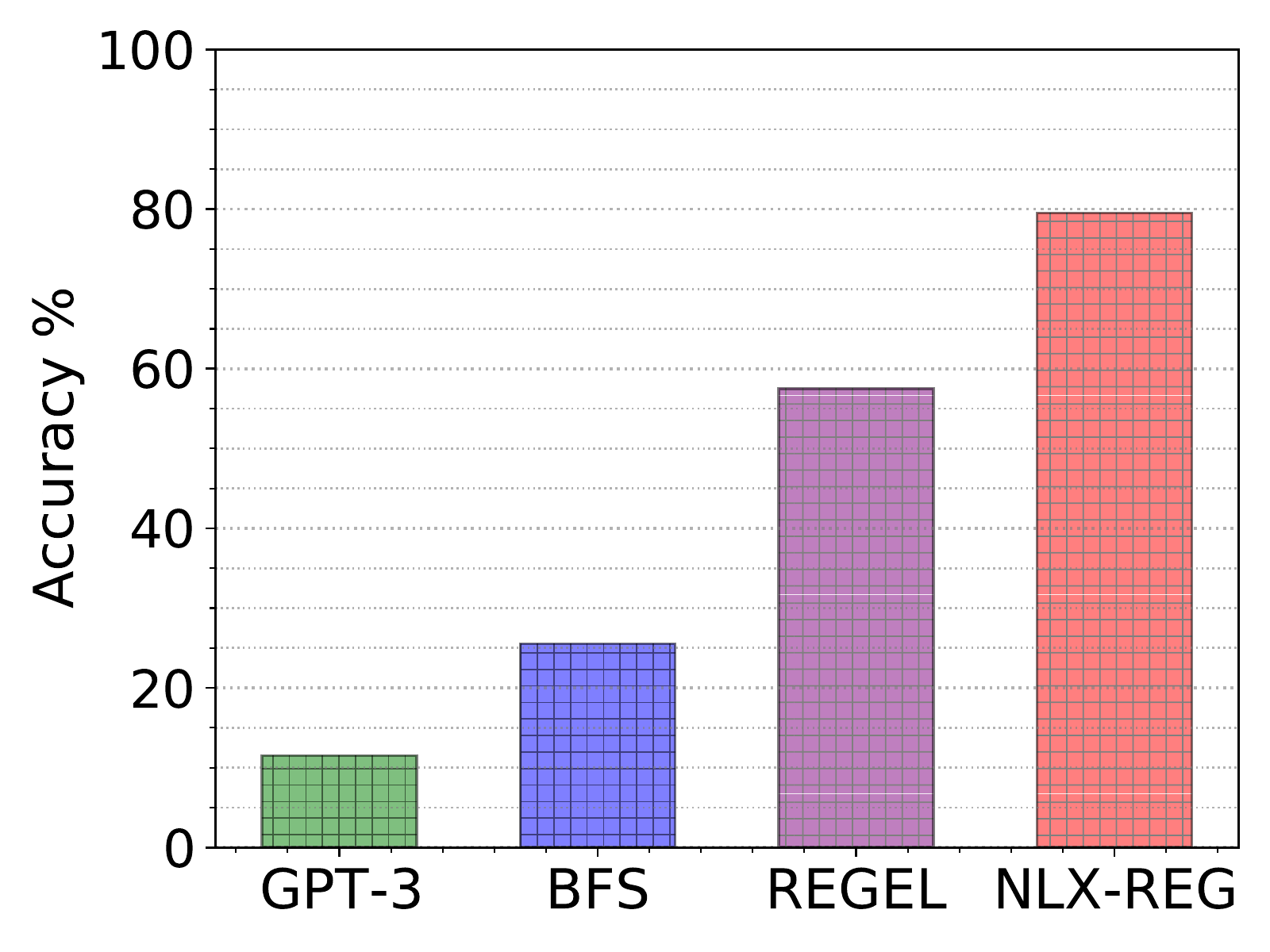}
\caption{Average}
\label{subfig:avg}
\end{subfigure}
\hfill
\vspace{-2mm}
\caption{Evaluation of \textsc{nlx-reg} and baseline systems}
\label{fig:reg_res}
\end{figure}

\subsubsection{Experimental Setup} 

For each benchmark set, we computed the PTM's prompt using a subset of tasks (question-answer pairs). For DeepRegex, as there was significant training data available we used 10 tasks for the prompt chosen from the training set as described in \autoref{sec:opt}. For StackOverflow, where there was significantly less training data, we used 5 out of the 25 tasks in the prompt. 
The remaining tasks were used for evaluation. Following \cite{regel}, each system was given 60 seconds for each task. A task is considered successfully done, if the output expression is \emph{semantically} equivalent to the ground-truth of that task. We used an off-the-shelf tool, RFixer~\cite{rfixer}, for deciding if two expressions are semantically equivalent. 
If the synthesized program, $p$, is not equivalent to the ground truth, $g$, the 
synthesizer under test is given another attempt with additional examples (up to 10 iterations).
 In particular, one negative example accepted by $p$ 
and rejected by $g$ (if it exists) and another positive example accepted by $g$ and rejected by $p$ (if it exists) are added to the task. We relied on RFixer to automatically generate such examples by  comparing $p$ and $g$ semantically. 
This procedure is in accordance with how real users interact with program synthesizers; once a candidate program is found,  the user either accepts it or provides additional examples to guide the synthesizer to find the intended program.


 
 \subsubsection{Comparison to the State-of-the-Art}

  \autoref{fig:reg_res} presents the accuracy of 
 \textsc{nlx-reg} and the
 baseline systems when applied on the DeepRegex (\ref{subfig:dr}) and  on the StackOverflow (\ref{subfig:so})
 data-sets. The average results across both data-sets is provided in \autoref{subfig:avg}.
 All systems performed considerably better on the DeepRegex data-set, where  tasks are relatively less complex than those in StackOverflow. 
 Both \textsc{nlx-reg} and REGEL achieve a high accuracy on the DeepRegex data-set by solving 104 ($90\%$) of the cases; while BFS and GPT-3 systems were less successful and only solved 49 ($41\%$) and 21 ($18\%$) of the cases. 
 On the StackOverflow data-set, however, 
 \textsc{nlx-reg} outperforms all baselines by solving 14 ($70\%$) cases. REGEL solved 5 ($25\%$) and BFS and GPT-3 solved respectively 2 ($10\%$) and 1 ($5\%$) cases on this data-set. 
 This shows the effectiveness of our approach in how, despite being domain-agnostic in nature, it was able to meet the performance of the specialized REGEL system with marginal difference on one dataset, and significantly outperform it on the other. Across both data-sets,
 \textsc{nlx-reg} solves $80\%$ of the tasks, 
which is $23\%$ better than the closest baseline, REGEL.
 
    \begin{wrapfigure}{r}{0.35\textwidth}
\includegraphics[width=0.33\textwidth]{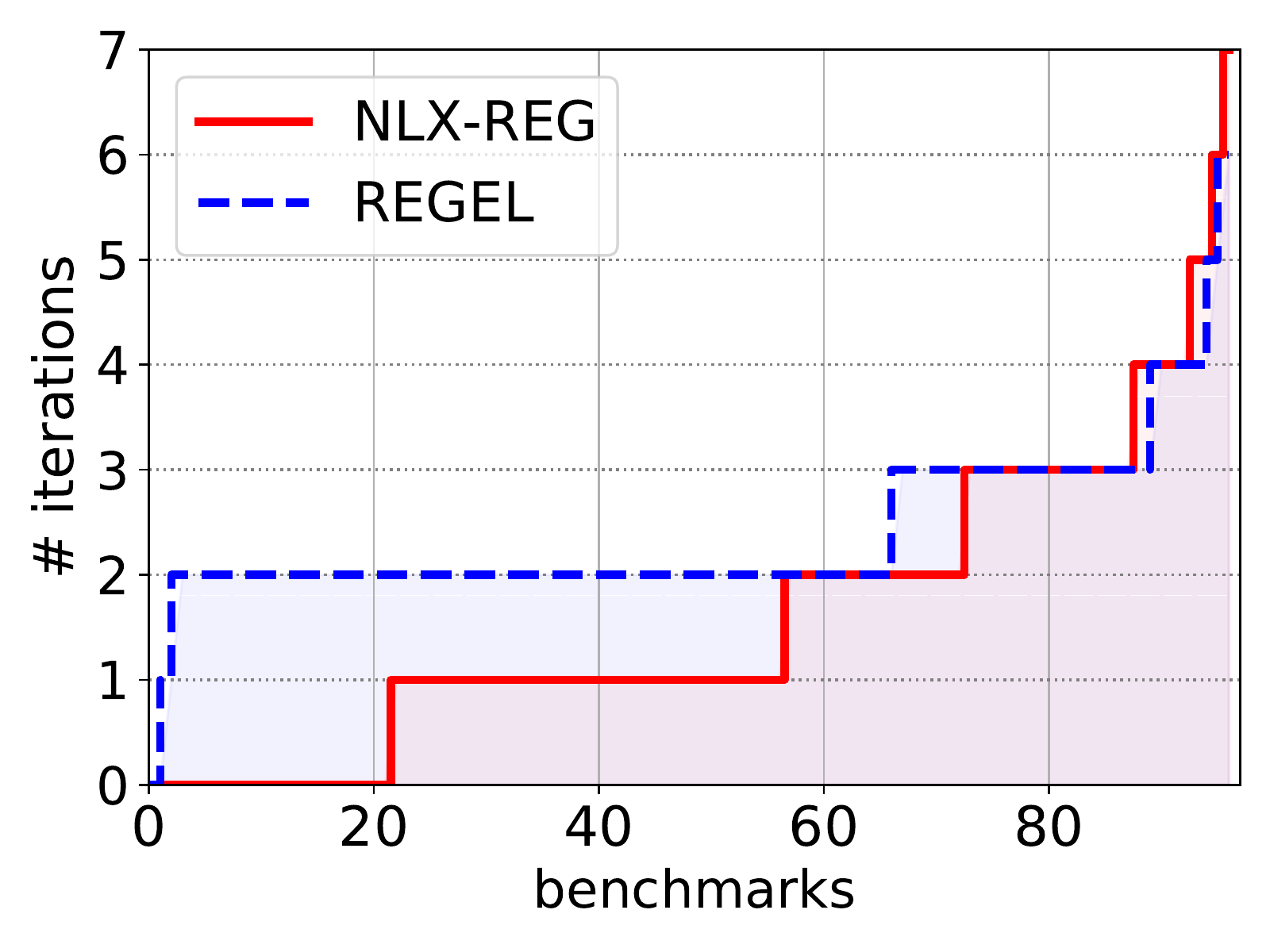}
\caption{Number of iterations before success}
\label{fig:iter}
\end{wrapfigure}

 To assess 
 how effectively each system leverages additional examples and converges to the ground-truth,  we performed further experiments on the subset of benchmarks that both
  \textsc{nlx-reg} and REGEL successfully solved, in order to compare the number of examples required by the two systems on tasks where both systems succeeded. \autoref{fig:iter} presents the results. The y-axis shows the number of iterations in which examples are provided before the correct program is obtained, with $0$ meaning that the synthesizer's first guess was correct and no additional examples were needed. The x-axis shows the number of benchmarks in each category. 
On average \textsc{nlx-reg} required 1.5 rounds and REGEL required 2.3 rounds of additional examples.
In particular, \textsc{nlx-reg} successfully guessed the ground-truth at first trial in 22 cases; this number for REGEL was 2. Similarly, 35 cases required only one round of additional examples for \textsc{nlx-reg}; this number was 1 for REGEL. Both systems require 3 or more additional rounds for about one third of all cases.

\REV{
\subsubsection{Ablation Study}
In addition to the comparison with the state-of-the-art described above, we also conducted an ablation study~\cite{DBLP:journals/corr/abs-1901-08644} with the goal of understanding the impact of the main components of the \textsc{nlx} technique on the overall performance of the system.
Specifically, we defined four versions of the system (v1-v4) where each version replaces a specific component of the algorithm with a naive alternative solution. We describe each of these versions below.}

\REV{
In system v1, we initialize the cache with \textit{all} atomic components found in \textit{any} of the PTM candidates, in order to assess the impact of our initialization procedure (discussed in §\ref{subsec:init}).
In v2 our expansion methodology (§\ref{subsec:expand}) is replaced with a full application of all DSL operators, i.e. expansion is done by a \textit{complete enumeration} of all valid terms in the DSL. In v3 our ranking procedure (§\ref{subsec:rank}) is replaced with a function that \textit{randomly} selects the final output of the system.  Lastly, v4 represents the full \textsc{nlx-reg} system but where we use a \emph{fixed} set of randomly-chosen cases for the prompt given to the PTM for each task (instead of dynamically choosing the prompt from the full training set that is available). This version is designed to assess the effectiveness of our prompt-generation technique (presented in~§\ref{sec:opt}) and also to evaluate the scenario where the user only has a very small amount of training data for the prompt.
}

\begin{figure}[t]
\begin{subfigure}{0.53  \textwidth}
\includegraphics[width=\textwidth]{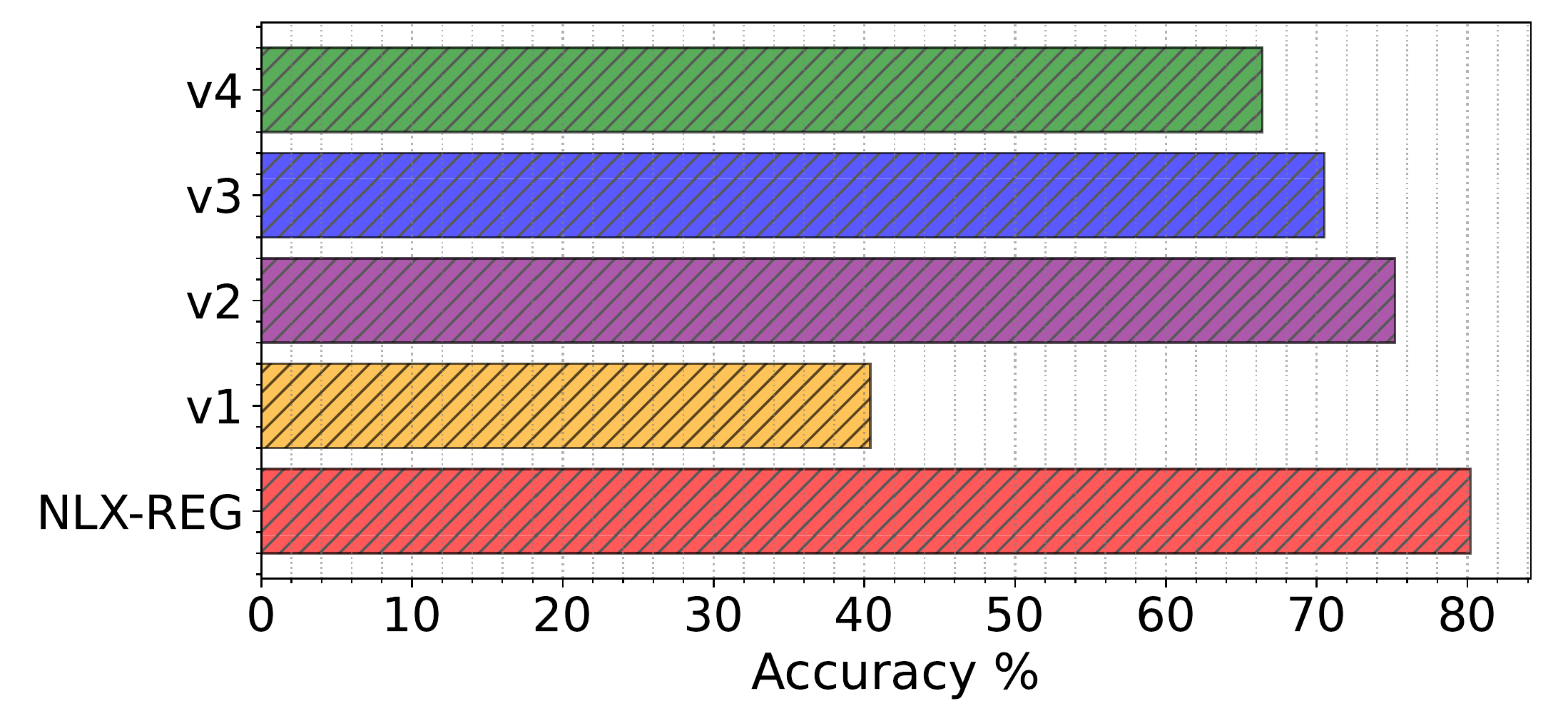}
\caption{performance comparison between different versions}
\label{subfig:abl1}
\end{subfigure}
\hfill
\begin{subfigure}{0.46\textwidth}
\center
\includegraphics[width=0.7\textwidth]{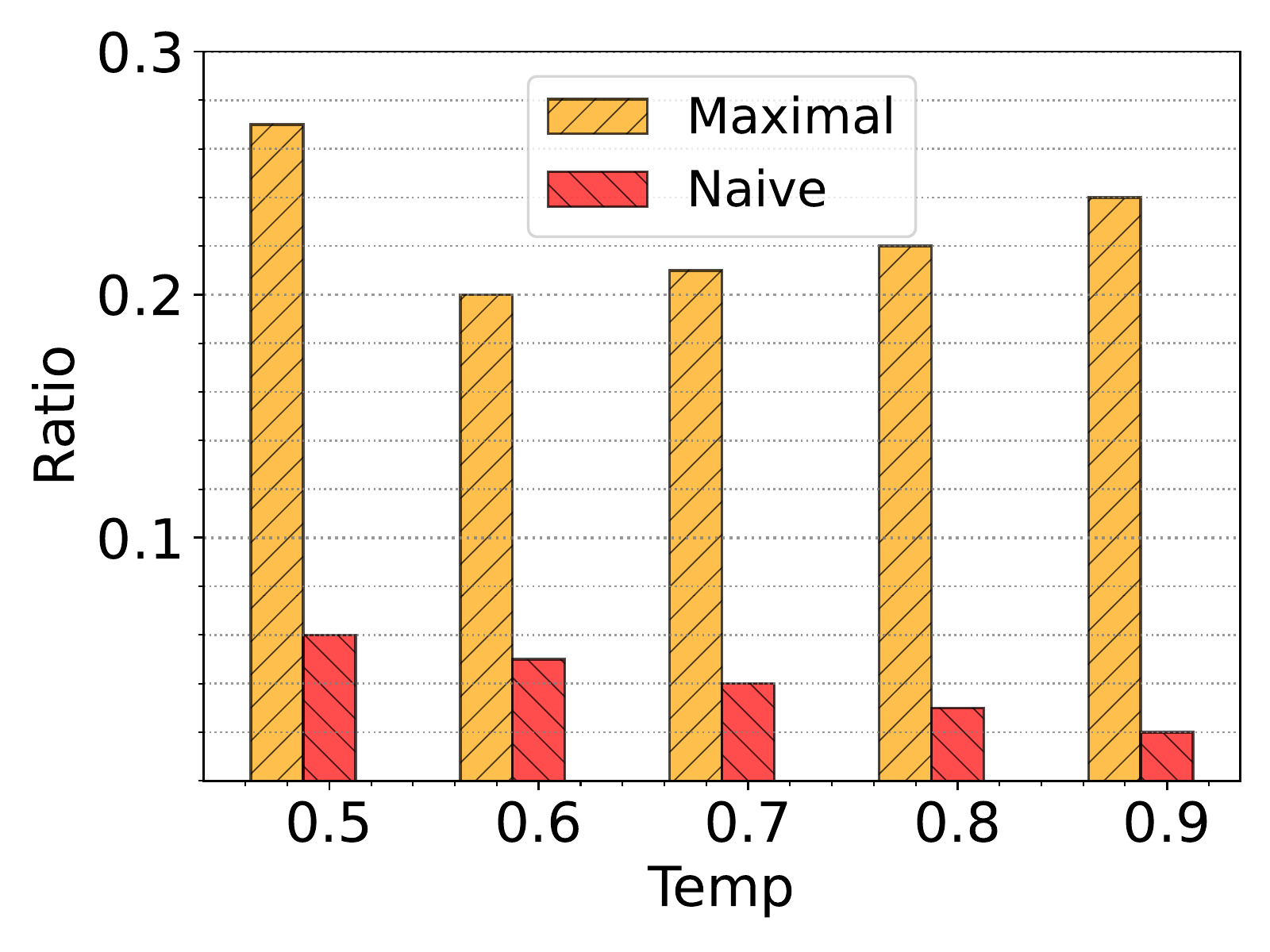}
\caption{ratio of terms appearing in the ground-truth}
\label{subfig:abl2}
\end{subfigure}
\caption{Ablation Study of \textsc{nlx-reg}}
\label{fig:ablation}
\end{figure}

\REV{\autoref{subfig:abl1} presents the results from our ablation study. The y-axis is labeled with the versions of the system defined above and the x-axis represents the average success rate across both data-sets. Firstly, we note that all four versions of the system perform more poorly than the full system: specifically, v1 through v4 achieve 40\%, 5\%, 10\% and 14\% lower success rates than the full \textsc{nlx-reg} system respectively. This shows how each of the component phases of the \textsc{nlx} technique contributes to the overall performance gain of the \textsc{nlx-reg} system compared to the BFS version defined earlier (the BFS setup can in fact be  thought of as an amalgamation of v1, v2 and v3). 
}

\REV{
We also observe that the most significant degradation of 40\% is seen in v1, which highlights the importance of the core initialization phase that uses our maximal components technique. We delved further into this to analyse the \emph{quality} of the PTM's outputs by measuring the ratio of terms appearing in any PTM candidates which also appear in the ground-truth program. \autoref{subfig:abl2} presents this analysis for different temperature settings of the GPT-3 model, where temperature is a parameter of the PTM that controls how much randomness occurs in the PTM output. We observe that when considering \emph{all components} from the PTM output, this ratio is generally very small (less than 0.06) for all temperatures and gets smaller as the temperature increases and more randomness occurs in the PTM output. This shows that a large majority of the terms appearing in the PTM's output are redundant and can potentially harm the synthesis procedure. As discussed in §\ref{subsec:init}, our initialization technique based on maximal components is designed to address this issue. \autoref{subfig:abl2} shows that  repeating this experiment but only counting components obtained using our initialization technique, the ratio increases significantly (about 8X) compared to the naive approach. Moreover, there is also not a significant relative decline in the ratio as temperature increases. This illustrates how our initialization procedure effectively declutters the output of the PTM to enhance the overall performance of the system and is also robust to temperature variations that may introduce more randomness.}

\REV{Finally, we note that the system v4 is not only an experimental instantation for ablation evaluation purposes, but represents the very realistic practical scenario where the user only has a very small amount of training data available (like 10 example pairs only): in such cases our prompt-generation technique of ~§\ref{sec:opt} is not applicable as the user can simply provide all the data they have. We observe that while v4 has a significant decrease of 14\% compared to the full system, the overall accuracy of v4 is 66\%, which is still significantly better than the overall accuracy of the REGEL system which is 57.5\% despite it being trained on much more data. This shows that the benefit of the \emph{few-shot learning} capability of PTMs is also exhibited by our multi-modal \textsc{nlx} system because with only a handful of examples it can still perform better than the state-of-the-art system that has been trained on much more data. But of course, our prompt-generation techniques provide significant further gains in the situations where we do have all the training data available.
}

%

\subsection{Domain of CSS Selectors}
\label{subsec:css}


\REV{Although the main focus in this work is the domain of regular expressions, a notable distinguishing characteristic of our approach is that it is  domain-agnostic in nature. This is because the techniques are not designed specifically for the language of regular expressions and can in theory be applicable to other DSLs. Though it is not an extensive exploration of applicability to arbitrary languages, we evaluate the generality aspect of our approach by  performing a preliminary evaluation of an implementation of our algorithm in the very different domain of CSS selectors (Cascading Style Sheets) \cite{css}.} CSS selectors are expressions for selecting elements from the document object model (DOM) of a webpage, based on structural properties that are defined by the HTML source markup of the webpage. We use the language of CSS selectors shown in Figure \ref{fig:cssdsl}.  

\textbf{Dataset.} We collected real-world scenarios from questions about CSS selectors posted on StackOverflow. We searched for such questions using the tags "css" and "css-selectors", and as in the previous section, created concise natural language descriptions for the selector based on the description in the question. Some examples of such tasks are shown in Figure \ref{fig:cssexamples}. Out of 25 such cases we excluded 6 that were using \emph{pseudo-classes} such as $\mathtt{:hover}$ or $\mathtt{:focus}$, which are not static properties of the input webpage and not handled by our CSS parser. This left a total of 19 cases in the dataset. For each of these tasks in this dataset, we also needed a sample input webpage on which one can execute and test the selectors and provide examples of desired elements that should be selected. We synthetically created such a sample webpage by manually examining each of the selectors in the dataset and creating representative HTML structures that contain positive and negative examples for each of the selectors.

\input{Figs/CSS}

\textbf{System and baselines} We implemented our system $\textsc{nlx-css}$ for multimodal synthesis of CSS selectors as an instantiation of our generic algorithm from Figure \ref{fig:alg} for the CSS domain. The DSL we used was $\lang_{\textsc{css}}$ from Figure \ref{fig:cssdsl} and the language model $\model_{\textsc{css}}$ was obtained using GPT-3 with few-shot training for the CSS domain. Given the small size of our dataset of only 19 cases, we used 3 of these for the few-shot training examples for GPT-3, and the remaining 16 cases were used as the test set. As we had chosen CSS as a novel domain of study, we are not aware of prior work for multi-modal synthesis of CSS selectors. Hence  the two baselines we chose were GPT-3 by itself (the top-ranked program from the PTM model $\model_{\textsc{css}}$ given only the natural language query), and the brute-force multi-modal approach $\textsc{BFS}$ which represents an enumerative search starting from all atomic components of the top-ranked programs from the PTM model $\model_{\textsc{css}}$.

\textbf{Evaluation} We evaluated our system and the two baselines using the test dataset of 16 cases. As in the regex domain, for the multi-modal systems we provided examples iteratively in a CEGIS fashion for a maximum of 10 iterations. A task was considered successfully completed if the synthesized selector is semantically equivalent to the ground truth selector given for that task. As there was no automated equivalence checker for this domain, we performed the equivalence check by manual inspection at every iteration. At each iteration, if the synthesizer under test did not produce the correct selector, then another positive and negative example element was provided from our sample webpage.

\begin{wrapfigure}{r}{0.36\textwidth}
\includegraphics[width=0.32\textwidth]{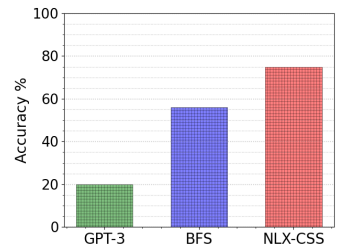}
\caption{Evaluation of \textsc{nlx-css}}
\label{fig:csseval}
\end{wrapfigure}

The results of our experiments are shown in Figure \ref{fig:csseval}. The relative performance of the three systems are similar to the previous section, with our system $\textsc{nlx-css}$ performing the best with 75\% accuracy, the brute force approach at 56\% and GPT-3 at 20\%. As in the regex domain, we observed the benefits of our approach in obtaining relevant components from the PTM candidates and guiding the search based on similarity to these programs. For example, for case I in Figure \ref{fig:cssexamples} the initial components included the composite expression $\mathtt{input[type="checkbox"][value]}$ which required minor repairs to construct the correct program. In cases II and III we observe the similarity of the operators used in the ground truth and the PTM candidates, even though none of the PTM results were exactly equivalent to the ground truth. 

As for the number of examples required by our system to successfully address the task: the average number of examples iterations required to return the correct program was 1.6, with only 2 cases requiring more than 2 iterations.

This is a preliminary evaluation mainly due to the small size of the dataset and the manual work such as equivalence-checking required for experimentation. In particular, using only 3 examples for the few-shot prompt training of GPT-3 was a notable limitation, and we can expect improved performance of all systems with more prompt training examples for GPT-3. This is evident from an examination of the failure cases where the main reason for failure was that the GPT-3 candidates were very different from the ground-truth programs in these cases (often including non-CSS syntax) which meant that many relevant components/operators required for synthesis were missing in these cases. However, while the accuracy of the underlying language model can improve arbitrarily with more prompt-training data or even fine-tuning, the key result of this study is the relative performance of the systems. It demonstrates the added benefit of the synthesis techniques to address the challenging cases that cannot be directly handled by the language model and require further interaction with examples.   

%% file: Figs/CSS.tex
\begin{figure*}[t]
\footnotesize
    \centering
    \scalebox{0.91}{%
    \begin{tabular}{ | c | l | l | l |}
    \hline
    \textbf{\#} &
     \multicolumn{1}{|c|}{\textbf{Natural Language}} & \multicolumn{1}{|c|}{\textbf{Ground Truth}} & 
     \multicolumn{1}{|c|}{\textbf{ Pre-trained Model's Candidates}}
      \\ \hline \hline
    \textsc{i} &
    $
    \begin{array}{l}
    \text{select just those checkboxes}\\
    \text{that have values set}\\
    \end{array}
    $
    &
    input[value][type="checkbox"]:not([value=""])
    &
    \hspace{-6mm}
    $
     \scriptsize
    \arraycolsep=21.6pt
    \begin{array}{l}
       \text{ input[type="checkbox"][value]} \\ 
       \text{ [type="checkbox"][value]} \\ 
       \text{ [checked="true"] } \\ 
       \text{ input[type="checkbox"]:checked} \\ 
       \text{ [value]} \\ 
       
    \end{array}
    $
    \\ \hline
    \textsc{ii} &
    $
    \begin{array}{l}
    \text{something that matches}\\
    \text{"(.a or .b) and .c"}\\
    \end{array}
    $
    &
    .a.c,.b.c
    &
    \hspace{-6mm}
    $
     \scriptsize
    \arraycolsep=21.6pt
    \begin{array}{l}
        \text{ .a+.b+.c } \\ 
        \text{ .a.b.c } \\
        \text{  a.b[class*="c"] } \\
        \text{ [class~="a"][class~="b"] .c } \\
        \text{ .a|.b[.c] } \\
    \end{array}
    $
    \\ \hline
    \textsc{iii} & 
    $
    \begin{array}{l}
        \text{select the first and the} \\
        \text{last TD in a row}
    \end{array}
    $
    &
        \text{tr td:first-child, tr td:last-child}
    & 
    \hspace{-2mm}
    $
    \scriptsize
    \arraycolsep=11.6pt
    \begin{array}{l}
        \text{tr:first-child:last-child} \\
        \text{td[last()]}
    \\
    \text{td:first-child:last-child} \\ \text{td:first-child,a:td:last-child}
    \\
    \text{td:not(:first-child):not(:last-child)} \\ \text{td:nth-child(1), a:td:last-child}
    \end{array}
    $

 \\
 \hline
    \end{tabular}
    }
    \caption{Example tasks for inferring CSS selectors. For each task we show the natural language description, the desired ground truth selector, and a sample of the PTM's top-ranked programs }
    \label{fig:cssexamples}
\end{figure*}

%% file: related.tex
\section{related work}
\label{sec:related}

\ourpara{Regex Synthesis:}
There is a large body of prior work on synthesizing regular expressions from examples.
Angluin presented algorithms to learn finite state automata and regular expressions from a given set of positive and negative examples~\cite{angluin1978, angluin1987}.
Recently, Mina et.al revisited the problem of learning regular expressions from introductory automata assignments using both positive and negative examples, which leverages ideas of over and under approximation to reduce the search space~\cite{mina2017}.
Since it is hard to construct regular expressions correctly, there is also work on the problem of repairing regular expressions to help developers.
Given an incorrect regular expression and a set of positive and negative examples, RFixer returns the closest regex (to the original one) that satisfies the examples~\cite{rfixer}.
Our system is different in that it takes both NL and examples to find the intended program with much fewer interactions.

Researchers have been looking at natural language as a specification to generate regular expressions. Kushman et. al. proposed a semantic parser to synthesize regular expressions from natural language descriptions~\cite{kushman2013}.
With the rise of deep learning, there is recent work that formulates this problem as a machine translation problem (seq2seq) to translate descriptions to regular expressions~\cite{deepregex, semregex}.
Unlike these systems, we also allow examples in addition to natural language to refine the intent. 

Finally, there is some recent work on regular expression synthesis using a combination of natural language and examples, which we will discuss below.

\ourpara{Multimodal Synthesis:}
Because different specification modalities have different characteristics (e.g., natural language description is versatile but ambiguous while examples are sound but incomplete), recent work on \emph{multi-modal synthesis} has been leveraging combinations of multiple types of specifications.
Manshadi et. al. discussed a probabilistic PBE system to perform string transformation using examples and natural language~\cite{manshadi2013}.
This work extended the version-space-algebra in~\cite{flashfill} by allowing the edges to carry probabilities calculated from both program properties and natural language descriptions.
Raza et. al. presented a multi-modal synthesis system that first maps descriptions into various concepts and uses the examples to refine the concepts~\cite{raza2015}.
\texttt{MARS} is a system that synthesizes data wrangling operations from a combination of input-output examples, natural language description, and partial code snippets~\cite{chen2019maximal}.
Their technique uses a combination of sequence to sequence (seq2seq) model to maps the description to an abstract program (sketch) and the apriori algorithm to mine the association rules. 
The entire problem was then reduced to a Max-SMT problem. 

\ourpara{Multimodal Regex Synthesis:}
Recently, there has been some work on synthesis of regular expressions from a natural language description and input-output examples~\cite{regel,DBLP:journals/corr/abs-2012-15489}.
The Regel system ~\cite{regel} first uses a semantic parser to parse a description into a sketch, then completes the sketch using enumerative search guided by the examples.
 In contrast, we utilize a PTM to generate components from the description and then use a novel CBS synthesis algorithm, which is guided by the PTM output, to generate a program that satisfies the examples.
We show higher accuracy of our technique on real-world  benchmarks in comparison to the Regel system. Furthermore, while Regel is designed specifically for the regular expression domain, our approach is domain-agnostic and applicable to other programming domains.

\REV{Another recent related work in this area is the TransRegex system ~\cite{DBLP:journals/corr/abs-2012-15489}. TransRegex is based on two distinct phases of first generating a best regex using an NL model, and then an independent examples-based repair technique to repair this best regex.  While we were unable to find an implementation of this system for direct evaluation, their reported accuracy on realistic scenarios from Stack Overflow reaches towards 70\% which is similar to our results. The key difference again is that this system is highly specialized for the Regex domain, while our approach is domain-agnostic and  applicable in at least one other domain of CSS. In terms of technique, in contrast to the two distinct-phase approach of TransRegex, our approach more tightly integrates synthesis and NL by using the \textit{set} of top candidate programs to guide the synthesis at multiple stages (initialization, expansion and ranking).  This tight integration has the benefits that it can "mix and match" likely components that may not all necessarily occur in the top program, and can also analyse patterns of operator occurrences across the top programs to infer the overall shape of the target program. We also note that the major contributions of~\cite{DBLP:journals/corr/abs-2012-15489} are around training of the model to reward syntactically valid regular expressions and to bake in semantic equivalence of regular expressions -- these steps are orthogonal to our work and may even be applied as domain-specific optimizations on the output of GPT-3 to improve our system further. 
}

\ourpara{Enumerative Synthesis:}
Enumerative search is one of the simplest program synthesis techniques, yet is proven to be effective for synthesizing small programs in complex search space~\cite{SyGus, alur2015synthesis, eusolver}.
Alur et. al. formalized the syntax-guided synthesis (SyGus) problem (where the search space are programs in a CFG) and proposed three different instantiations of the counter-example-guided-inductive-synthesis
(CEGIS) strategy~\cite{SyGus}. 
Subsequently, \cite{alur2015synthesis} extends CEGIS with through unification, where the idea is to unify different programs that satisfy different parts of the inputs.
EUSolver makes the enumeration process more efficient by employing a divide-and-conquer approach~\cite{eusolver}.
In addition to techniques that are based on program size, researcher also proposed new search techniques such as abstraction-based~\cite{abstract-synthesis, component-table, refinement-type}, constraint-based~\cite{sketching, verification-synthesis, oracle-guided}, deep-learning-based\cite{deepcoder}.

Raza et. al. introduced \emph{predictive synthesis}, in which the synthesizer learns a data wrangling program from just the input (without the output example)~\cite{raza2017automated}. 
Their approach enumerates the program literals bottom-up and has a search strategy that biases conforming programs.
Some works have also looked at combining enumerative and deductive synthesis~\cite{raza2020web, enumerative-deductive}.
Our approach also employs enumerative synthesis, but instead of generating components from scratch, we employ a PTM to generate \emph{maximal} components and utilize a novel search technique to synthesize the final regex.

\REV{\ourpara{Closed frequent itemset mining:}
Our goal of finding the most valuable initial components has similarities to the field of frequent pattern mining in databases \cite{agrawal93}. Similar notions of frequency and redundancy are also considered in \emph{closed frequent item-set mining} \cite{phm00} where the goal is to find frequent sets while also avoiding redundancy by not finding subsets with the same support. However, the underlying focus in this field is on “association rules” that follow a flat set-based structure and exist in independent records of the database. The key conceptual difference in our case is that the entities of interest are not sets but structured AST components that may be nested inside one another, and where redundancy comes from the sub-tree rather than subset relation. 
}

\ourpara{Natural Language to Code:}
There have been numerous proposals to generate different kinds programs from natural language, including SQL queries~\cite{meta-learning-natural,sqlizer, rat-sql}, smartphone automation scripts~\cite{smartsynth}, spreadsheet formulas~\cite{nlyze}, bash~\cite{nl2bash}.
SQLizer generates a sketch from natural language, then refines it using probabilistic type inhabitation and automated sketch repair~\cite{sqlizer}.
SmartSynth combines semantic parser with type-based synthesis to generate phone automation scripts~\cite{smartsynth}.
NLyze introduced a translation algorithm that utilizes spatial and temporal context in the spreadsheet~\cite{nlyze}.
Recently, RAT-SQL tackles the NL to SQL problem by using relation-aware self-attention to incorporate reasoning that involves both question entities and database schema~\cite{rat-sql}.
Unlike the above techniques, our approach does not require explicit supervision with large curated datasets for particular domains, as it  leverages the strength of PTMs to provide robust coverage of various domains with few-shot learning.

\REV{
\ourpara{Natural Language to Code using PTMs:}
There has not been much published work on code generation using large language models, e.g. GPT-3. There is however plenty ongoing activity and we expect various techniques to emerge in near future. Codex~\cite{DBLP:journals/corr/abs-2107-03374} is GPT-3 fine-tuned on code, and generates Python code from docstrings in about 30\% of cases. In contrast, for restricted domains, and using synthesis as a post-processor for GPT-3 as described here, we are able to get much higher precision. Going forward, fine-tuned models can be used along with synthesis-based post-processing to build powerful NL to code systems.  \citet{DBLP:journals/corr/abs-2105-09938} introduced a large benchmark set for coding tasks (dubbed APPS), that can be used to systematically evaluate the ability of such techniques in using various data-structures and programming techniques. These benchmarks assume a general purpose language (Python) which is currently not what our NLX system is targeting. However, we think APPS is a valuable framework to track advancements in program synthesis research and would be interesting to explore in our future works.
}

%% file: future.tex
\section{Discussion and Future Work}

\textsc{NLX} is a general approach for multimodal synthesis that combines the strengths
of PTMs and program synthesis.  Its effectiveness is based on the underlying
hypotheses that 
(A) the multiple candidates returned by PTMs contain the components
of the ground truth, even though they may not contain the whole ground truth,
(B) furthermore, the candidates reveal (approximately) the distribution/frequency 
of the operators in the ground truth, 
(C) users can provide input-output examples to refine their intent in case the
synthesis engine does not return their desired program, and 
(D) subprograms (subterms) can be executed on inputs (obtained from the examples)
to determine (approximate) semantic equivalence of these subprograms.

While our experimental evaluation has focused on regular expressions and CSS selectors,
we have observed that some of these assumptions, specifically (A) and (B),
hold in general. There exist domains where assumptions (C) and particularly
(D) are not easily satisfied, but even in those cases, the \textsc{nlx} approach can 
be adapted by replacing the steps that rely on examples by alternate steps that 
rely on other forms of intent specification.

A particularly interesting domain is that of Python's data processing library
Pandas.  Pandas is popular among data scientists for writing scripts that can be used to 
ingest data, clean data, reshape and manipulate data, and visualize data.
Pandas is a good target for generating code from NL descriptions because 
(a) it is widely used, including by non-programmers, and 
(b) it has a very large API, and it is very difficult to remember the API details, especially
for an occassional user.

\input{Figs/Pandas}

We validated hypotheses (A) and (B) for the Pandas domain by collecting NL descriptions 
along with ground-truth expressions from stackoverflow posts about Pandas. 
We then used a PTM with dynamic prompt to generate 25 candidate programs for the NL descriptions. 
We then analyzed if the candidates have the components used in the ground-truth program.  
There were around 40\% benchmarks where the ground-truth program was present in the 25 candidates. 
Figure~\ref{fig:pandasexamples} shows three instances when the ground truth was not present in the candidates returned by PTM. There were three classes of instances.
In the first class, the candidates contained all components used in the ground truth.
For example, consider the ground-truth program 
$\text{df.groupby(['HP', 'Type 1']).agg(['mean', 'count'])}$ 
that corresponds to the NL description "group by HP and Type 1 and calculate mean and count for each group".
Among the generated candidates, we find the maximal components
$\text{df.groupby}$,
$\text{['HP', 'Type 1']}$,
$\text{agg}$, 
$\text{np.mean}$, and
$\text{'count'}$, which can be combined to give the program
$\text{df.groupby(['HP', 'Type 1']).agg([np.mean, 'count'])}$,
which is equivalent to the ground-truth program. 
The other two examples in Figure~\ref{fig:pandasexamples} show cases where the
candidates do not contain all the components needed to recreate the ground truth, but 
they contain many of the components. In the first row, only one component, namely
$\text{set\_index}$ is missing, whereas in the last row 
three are missing, namely $\text{to\_series}$, $\text{df.dtypes}$, and $\text{groups}$.  

A key difficulty in using \textsc{nlx} for synthesizing Pandas code is that 
assumptions (C) and (D) are harder to satisfy. In such cases, we need to adapt the approach 
and extend it with other methods, such as, the use of types to suggest repairs, 
which we leave for future work. We can also extend \textsc{nlx} by generalizing the notion of components to also 
include {\em{sketches}}, or terms with holes.  Most of the steps in our algorithm will
generalize to using sketches as components, except for steps that require
assumption (D). Adapting \textsc{nlx} to also use sketches as components is an interesting
direction for future work. It is also possible to consider fine-tuning the pre-trained models.  Fine tuning requires more
data, but it also provides more value by giving a good set of initial candidates
to the synthesis procedure.  There is also the future possibility of employing constrained decoding
to guarantee that the pre-trained model only generates valid code in the target language.

\REV{
While we have discussed applicability to various specialized programming domains, it is also an interesting question to ask if such techniques can be applicable to much more expressive general purpose programming languages such as Python, Java or C\#. In practice we do not expect our techniques to directly scale to large programs in such highly expressive languages. However, it is an interesting research direction to build upon our ideas here. For instance, initial experiments on small code snippets in C\#  suggest that a \emph{compositional} approach to multi-modal synthesis may be valuable: instead of just input-output examples, if the user can provide “traces” of examples over some pseudo-code in natural language then that may more strongly guide the system to scale to more complex programs. These will be interesting explorations for future work.}

%% file: Figs/Pandas.tex
\begin{figure}[t]
    \begin{footnotesize}
    \centering
    \scalebox{0.89}{%
    \begin{tabular}{ | c | l | l | p{6.5cm} |}
    \hline
    \textbf{\#} &
     \multicolumn{1}{|c|}{\textbf{Natural Language}} & \multicolumn{1}{|c|}{\textbf{Ground Truth}} & 
     \multicolumn{1}{|c|}{\textbf{ Pre-trained Model's Candidates}}
      \\ \hline \hline
    \textsc{i} &
    $
    \begin{array}{l}
        \text{Convert a dataframe to a}
        \\ 
        \text{dictionary, where row values}
        \\
        \text{of 'HP' are the keys and the}
        \\
        \text{other columns are the values}
    \end{array}
    $
    &
        $\begin{array}{l}
            \text{df.set\_index('HP')}\\ \;\;\text{.to\_dict(orient='index')}
        \end{array}$
    &
    $
    \scriptsize
    \begin{array}{l}
        \text{pd.to\_dict(df, orient='index')}
        \\
        \text{df.to\_dict(orient='index', prefix='HP')}
        \\
        \text{df.to\_dict(orient='index')}
        \\
        \text{df.to\_dict(orient='index', axis=1)}
        \\
        \text{pd.DataFrame(df2).to\_dict(orient='records')}
        \\
        \text{df.to\_dict(orient='index', axis=1)}
    \end{array}
    $
    \\ \hline
    \textsc{ii} & 
    $
    \begin{array}{l}
        \text{group by 'HP' and 'Type 1'}\\
        \text{and calculate mean and}\\
        \text{count for each group}
    \end{array}
    $
    &
        $\begin{array}{l}
            \text{df.groupby(['HP', 'Type 1'])}\\ \;\;\text{.agg(['mean', 'count'])}
        \end{array}$
    & 
    $
    \scriptsize
    \arraycolsep=3.6pt
    \begin{array}{l}
        \text{df.groupby('HP', 'Type 1').mean().agg(['count', 'sum']).fillna(0)}
        \\
        \text{df.groupby(['HP', 'Type 1']).mean()}
        \\
        \text{df.groupby('HP','Type 1').apply(np.mean)}
        \\
        \text{df.groupby('HP','Type 1').apply(np.count)}
    \end{array}
    $
     \\  \hline 
    \textsc{iii} &
    $
    \begin{array}{l}
        \text{get list of columns}\\
        \text{grouped by datatype}\\
        \text{of column}
    \end{array}
    $
    &
        $\begin{array}{l}
            \text{df.columns.to\_series()} \\ \;\;\text{.groupby(df.dtypes).groups}
        \end{array}$
    &
    \hspace{0.7mm}
    $
    \scriptsize
    \arraycolsep=0pt    
    \begin{array}{ll}
        \text{df.groupby(['dtype'])}
        \\
        \text{df['name'].groupby('type').count()}
        \\
        \text{Code: df.groupby('col').columns}
        \\
        \text{df.groupby('column\_type').columns.value\_counts()}
        \\
        \text{grouped\_cols = df.groupby('Type')}
    \end{array}
    $
 \\
 \hline
    \end{tabular}
    }
    \end{footnotesize}
    \caption{Pandas Examples: The NL description of task, the associated ground truth, and the
    candidates generated by the PTM model. The last column only shows a few selected candidates.}
    \label{fig:pandasexamples}
    \vspace{-3mm}
\end{figure}

%% file: conclusion.tex
\section{conclusions}
\label{sec:conclusions}

This paper presents a novel technique for synthesizing 
programs from natural language descriptions and examples. 
We introduce a domain-agnostic algorithm that leverages the ability of modern pre-trained language models to provide probability distributions over program components from ambiguous natural language descriptions, and uses them to guide a novel component-based approach for synthesis from examples.
We instantiated our algorithm for two programming domains -- the domains of regular expressions and CSS selectors. The experimental results suggest effectiveness
of this approach on both domains. Most notably, 
our domain-agnostic synthesizer when specialized to the domain of regular expressions outperforms the
state-of-the-art and highly-specialized synthesizer for this domain.

\ignore{
\begin{itemize}
    \item other domains
    \item sketches
    \item scalability. we have considered relatively small DSLs in this work, and remains to investigate how this approach will do for broader languages. We can expect to require stronger biases to maintain tractability of search
    \item better language models. With fine-tuning on top of pre-training we can expect to do even better in initial results (cite some work here)
    \item end with vision for ultimate goal we aim for:  to have a meta-synthesis system with which the user can easily create  a multi-modal synthesis system for any common DSL:  all you need to provide is the language (syntax and semantics) and a handful of training data for few-shot learning. 
\end{itemize}
}

%% file: app.tex
\section{Supplementary Definitions}
\label{app:def}
In this section we present the formal semantics of the DSLs introduced in the paper. \autoref{fig:regex-semantics} presents the semantics for regular expressions and \autoref{fig:css-semantics} presents the semantics for the domain of CSS selectors.

\begin{figure}[h]
\begin{footnotesize}
$
\begin{array}{lcl}
    \denote{\reg{i}}(s) = \bot &  & \mbox{ for all $\reg{i}$ in $\{\reg{0},\reg{1},\ldots\}$}
    \\
    \denote{\reg{c}}(s) = \bot &  & \mbox{ for all $\reg{c}$ in $\{\reg{A},\reg{B},\ldots\}$}
    \\
    \denote{\reg{fromChar(c)}}(s) = \top & \text{iff} & \mbox{$s==\reg{c}$}
    \\
    \denote{\reg{range(c_1,c_2)}}(s) = \top & \text{iff} & s==\reg{c} \mbox{ for some $c$ that lies between $\reg{c_1}$ and $\reg{c_2} 2$}
    \\
    \denote{\reg{union(s_1,s_2)}}(s) = \top & \text{iff} & 
      \denote{s_1}(s) = \top \mbox{ or }
      \denote{s_2}(s) = \top
    \\
    \denote{\reg{negate}(s)}(s) = \top & \text{iff} & \denote{s}(s) = \bot
    \\
    \denote{\reg{any}()}(s) = \top & &
    \\
    \denote{\reg{quant(e,i,j)}}(s) = \top & \text{iff} & 
      s = s_1s_2\ldots s_k,  i\leq k\leq j, \denote{e}(s_l) = \top \mbox{ for all $l\in\{1,\ldots,k\}$}
    \\
    \denote{\reg{quantMin(e,i)}}(s) = \top & \text{iff} & 
      s = s_1s_2\ldots s_j,  j \geq i, \denote{e}(s_k) = \top \mbox{ for all $k\in\{1,\ldots,j\}$}
    \\
    \denote{\reg{alter(e_1,e_2)}}(s) = \top & \text{iff} & 
      \denote{e_1}(s) = \top \mbox{ or } \denote{e_2}(s) = \top
    \\
    \denote{\reg{concat(e_1,e_2)}}(s) = \top & \text{iff} & 
      s = s_1s_2,  \denote{e_1}(s_1) = \denote{e_2}(s_2) = \top
    \\
    \denote{\reg{fromCharSet(s)}}(s) = \top & \text{iff} & \denote{s}(s) = \top 
\end{array}
$
\end{footnotesize}
  \caption{The semantics of regular expressions DSL}
  \label{fig:regex-semantics}
\end{figure}

\begin{figure}[h]
\begin{footnotesize}
$
\begin{array}{lcl}
    \denote{\css{i}} & = & \{ i \} \mbox{ for all number literals $i$}
    \\
    \denote{\css{MultipleOffset(i,j)}} & = & \{ j, i+j, 2i+j, 3i+j, \ldots \}  \mbox{}
    \\
    \denote{\css{s}} & = & s \mbox{ for all string literals $s$}
    \\
    \denote{\css{Any()}} & = & \mbox{the set of all nodes in the input document}
    \\
    \denote{\css{Union(n_1,n_2)}} & = & \denote{\css{n_1}}\cup\denote{\css{n_2}}
    \\
    \denote{\css{Not(n_1,n_2)}} & = & \denote{\css{n_1}} - \denote{\css{n_2}} \mbox{ where $-$ denotes set difference}
    \\
    \denote{\css{TagEquals(n, s)}} &= & \{ \mathtt{node} \in \denote{\css{n}} \mid \mbox{ the tag of $\mathtt{node}$ is "s"} \}
    \\
    \denote{\css{nthChild}(n, i)} & = & \{ \mathtt{node} \in \denote{\css{n}} \mid \mbox{ $\mathtt{node}$ is the $k$-th child of its parent for some $k$ in $\denote{\css{i}}$} \}
    \\
    \denote{\css{nthLastChild}(n, i)} & = & \{ \mathtt{node} \in \denote{\css{n}} \mid \mbox{ $\mathtt{node}$ is the $k$-th child for $k\in\denote{\css{i}}$, counting from the end, of its parent}\}
    \\
    \denote{\css{AttributeEquals(n, s_1, s_2}} & = & \{ \mathtt{node}\in\denote{\css{n}} \mid \mbox{ $\mathtt{node}$ has an attribute $s_1$ that is set to $s_2$} \}
    \\
    \denote{\css{AttributeContains(n, s_1, s_2}} & = & \{ \mathtt{node}\in\denote{\css{n}} \mid \mbox{ $\mathtt{node}$ has an attribute $s_1$ whose value contains $s_2$ as a substring} \}
    \\
    \denote{\css{AttributeStartsWith(n, s_1, s_2}} & = & \{ \mathtt{node}\in\denote{\css{n}} \mid \mbox{ $\mathtt{node}$ has an attribute $s_1$ whose value starts with $s_2$} \}
    \\
    \denote{\css{AttributeEndsWith(n, s_1, s_2}} & = & \{ \mathtt{node}\in\denote{\css{n}} \mid \mbox{ $\mathtt{node}$ has an attribute $s_1$ whose value ends with $s_2$} \}
    \\
    \denote{\css{RightSibling(n_1,n_2)}} & = & \{ \mathtt{node}\in \denote{\css{n_2}} \mid \mbox{ $\mathtt{node}$ is preceded by some node in $\denote{\css{n_1}}$ with the same parent} \}
    \\
    \denote{\css{Children(n_1,n_2)}} & = & \{ \mathtt{node}\in \denote{\css{n_2}} \mid \mbox{ $\mathtt{node}$ is the child of some node in $\denote{\css{n_1}}$} \}
    \\
    \denote{\css{Descendants(n_1,n_2)}} & = & \{ \mathtt{node}\in \denote{\css{n_2}} \mid \mbox{ $\mathtt{node}$ is the descendant of some node in $\denote{\css{n_1}}$} \}
\end{array}
$
\end{footnotesize}
  \caption{The semantics of CSS expressions DSL}
  \label{fig:css-semantics}
\end{figure}

%% file: main.bbl

\begin{thebibliography}{45}


\ifx \showCODEN    \undefined \def \showCODEN     #1{\unskip}     \fi
\ifx \showDOI      \undefined \def \showDOI       #1{#1}\fi
\ifx \showISBNx    \undefined \def \showISBNx     #1{\unskip}     \fi
\ifx \showISBNxiii \undefined \def \showISBNxiii  #1{\unskip}     \fi
\ifx \showISSN     \undefined \def \showISSN      #1{\unskip}     \fi
\ifx \showLCCN     \undefined \def \showLCCN      #1{\unskip}     \fi
\ifx \shownote     \undefined \def \shownote      #1{#1}          \fi
\ifx \showarticletitle \undefined \def \showarticletitle #1{#1}   \fi
\ifx \showURL      \undefined \def \showURL       {\relax}        \fi
\providecommand\bibfield[2]{#2}
\providecommand\bibinfo[2]{#2}
\providecommand\natexlab[1]{#1}
\providecommand\showeprint[2][]{arXiv:#2}

\bibitem[\protect\citeauthoryear{Agrawal, Imielinski, and Swami}{Agrawal
  et~al\mbox{.}}{1993}]%
        {agrawal93}
\bibfield{author}{\bibinfo{person}{R. Agrawal}, \bibinfo{person}{T.
  Imielinski}, {and} \bibinfo{person}{A. Swami}.}
  \bibinfo{year}{1993}\natexlab{}.
\newblock \showarticletitle{Mining Association Rules Between Sets of Items in
  Large Databases}. In \bibinfo{booktitle}{\emph{Proceedings of the 1993 ACM
  SIGMOD International Conference on Management of Data}}
  \emph{(\bibinfo{series}{SIGMOD '93})}, Vol.~\bibinfo{volume}{22}.
  \bibinfo{publisher}{ACM}, \bibinfo{address}{New York, NY, USA},
  \bibinfo{pages}{207--216}.
\newblock
\showISBNx{0-89791-592-5}
\urldef\tempurl%
\url{https://doi.org/10.1145/170035.170072}
\showDOI{\tempurl}


\bibitem[\protect\citeauthoryear{{Alur}, {Bodik}, {Juniwal}, {Martin},
  {Raghothaman}, {Seshia}, {Singh}, {Solar-Lezama}, {Torlak}, and
  {Udupa}}{{Alur} et~al\mbox{.}}{2013}]%
        {SyGus}
\bibfield{author}{\bibinfo{person}{R. {Alur}}, \bibinfo{person}{R. {Bodik}},
  \bibinfo{person}{G. {Juniwal}}, \bibinfo{person}{M.~M.~K. {Martin}},
  \bibinfo{person}{M. {Raghothaman}}, \bibinfo{person}{S.~A. {Seshia}},
  \bibinfo{person}{R. {Singh}}, \bibinfo{person}{A. {Solar-Lezama}},
  \bibinfo{person}{E. {Torlak}}, {and} \bibinfo{person}{A. {Udupa}}.}
  \bibinfo{year}{2013}\natexlab{}.
\newblock \showarticletitle{Syntax-guided synthesis}. In
  \bibinfo{booktitle}{\emph{2013 Formal Methods in Computer-Aided Design}}.
  \bibinfo{pages}{1--8}.
\newblock
\urldef\tempurl%
\url{https://doi.org/10.1109/FMCAD.2013.6679385}
\showDOI{\tempurl}


\bibitem[\protect\citeauthoryear{Alur, Cerny, and Radhakrishna}{Alur
  et~al\mbox{.}}{2015}]%
        {alur2015synthesis}
\bibfield{author}{\bibinfo{person}{Rajeev Alur}, \bibinfo{person}{Pavol Cerny},
  {and} \bibinfo{person}{Arjun Radhakrishna}.} \bibinfo{year}{2015}\natexlab{}.
\newblock \showarticletitle{Synthesis Through Unification}. In
  \bibinfo{booktitle}{\emph{Computer Aided Verification (CAV)}}.
\newblock
\urldef\tempurl%
\url{https://www.microsoft.com/en-us/research/publication/synthesis-through-unification/}
\showURL{%
\tempurl}


\bibitem[\protect\citeauthoryear{Alur, Radhakrishna, and Udupa}{Alur
  et~al\mbox{.}}{2017}]%
        {eusolver}
\bibfield{author}{\bibinfo{person}{Rajeev Alur}, \bibinfo{person}{Arjun
  Radhakrishna}, {and} \bibinfo{person}{Abhishek Udupa}.}
  \bibinfo{year}{2017}\natexlab{}.
\newblock \showarticletitle{Scaling Enumerative Program Synthesis via Divide
  and Conquer}. In \bibinfo{booktitle}{\emph{Tools and Algorithms for the
  Construction and Analysis of Systems}},
  \bibfield{editor}{\bibinfo{person}{Axel Legay} {and} \bibinfo{person}{Tiziana
  Margaria}} (Eds.). \bibinfo{publisher}{Springer Berlin Heidelberg},
  \bibinfo{address}{Berlin, Heidelberg}, \bibinfo{pages}{319--336}.
\newblock
\showISBNx{978-3-662-54577-5}


\bibitem[\protect\citeauthoryear{Angluin}{Angluin}{1978}]%
        {angluin1978}
\bibfield{author}{\bibinfo{person}{Dana Angluin}.}
  \bibinfo{year}{1978}\natexlab{}.
\newblock \showarticletitle{On the complexity of minimum inference of regular
  sets}.
\newblock \bibinfo{journal}{\emph{Information and Control}}
  \bibinfo{volume}{39}, \bibinfo{number}{3} (\bibinfo{year}{1978}),
  \bibinfo{pages}{337--350}.
\newblock
\showISSN{0019-9958}
\urldef\tempurl%
\url{https://doi.org/10.1016/S0019-9958(78)90683-6}
\showDOI{\tempurl}


\bibitem[\protect\citeauthoryear{Angluin}{Angluin}{1987}]%
        {angluin1987}
\bibfield{author}{\bibinfo{person}{Dana Angluin}.}
  \bibinfo{year}{1987}\natexlab{}.
\newblock \showarticletitle{Learning Regular Sets from Queries and
  Counterexamples}.
\newblock \bibinfo{journal}{\emph{Inf. Comput.}} \bibinfo{volume}{75},
  \bibinfo{number}{2} (\bibinfo{date}{Nov.} \bibinfo{year}{1987}),
  \bibinfo{pages}{87–106}.
\newblock
\showISSN{0890-5401}
\urldef\tempurl%
\url{https://doi.org/10.1016/0890-5401(87)90052-6}
\showDOI{\tempurl}


\bibitem[\protect\citeauthoryear{Balog, Gaunt, Brockschmidt, Nowozin, and
  Tarlow}{Balog et~al\mbox{.}}{2017}]%
        {deepcoder}
\bibfield{author}{\bibinfo{person}{Matej Balog}, \bibinfo{person}{Alexander
  Gaunt}, \bibinfo{person}{Marc Brockschmidt}, \bibinfo{person}{Sebastian
  Nowozin}, {and} \bibinfo{person}{Daniel Tarlow}.}
  \bibinfo{year}{2017}\natexlab{}.
\newblock \showarticletitle{DeepCoder: Learning to Write Programs}. In
  \bibinfo{booktitle}{\emph{Proceedings of ICLR'17}
  (\bibinfo{edition}{proceedings of iclr'17} ed.)}.
\newblock
\urldef\tempurl%
\url{https://www.microsoft.com/en-us/research/publication/deepcoder-learning-write-programs/}
\showURL{%
\tempurl}


\bibitem[\protect\citeauthoryear{Black}{Black}{1999}]%
        {lev}
\bibfield{author}{\bibinfo{person}{Paul~E. Black}.}
  \bibinfo{year}{1999}\natexlab{}.
\newblock \bibinfo{title}{Dictionary of Algorithms and Data Structures
  [online]}.
\newblock
\newblock
\urldef\tempurl%
\url{https://www.nist.gov/dads/HTML/Levenshtein.html (Accessed , March 2021)}
\showURL{%
\tempurl}


\bibitem[\protect\citeauthoryear{Brown, Mann, Ryder, Subbiah, Kaplan, Dhariwal,
  Neelakantan, Shyam, Sastry, Askell, Agarwal, Herbert-Voss, Krueger, Henighan,
  Child, Ramesh, Ziegler, Wu, Winter, Hesse, Chen, Sigler, Litwin, Gray, Chess,
  Clark, Berner, McCandlish, Radford, Sutskever, and Amodei}{Brown
  et~al\mbox{.}}{2020}]%
        {brown2020language}
\bibfield{author}{\bibinfo{person}{Tom Brown}, \bibinfo{person}{Benjamin Mann},
  \bibinfo{person}{Nick Ryder}, \bibinfo{person}{Melanie Subbiah},
  \bibinfo{person}{Jared~D Kaplan}, \bibinfo{person}{Prafulla Dhariwal},
  \bibinfo{person}{Arvind Neelakantan}, \bibinfo{person}{Pranav Shyam},
  \bibinfo{person}{Girish Sastry}, \bibinfo{person}{Amanda Askell},
  \bibinfo{person}{Sandhini Agarwal}, \bibinfo{person}{Ariel Herbert-Voss},
  \bibinfo{person}{Gretchen Krueger}, \bibinfo{person}{Tom Henighan},
  \bibinfo{person}{Rewon Child}, \bibinfo{person}{Aditya Ramesh},
  \bibinfo{person}{Daniel Ziegler}, \bibinfo{person}{Jeffrey Wu},
  \bibinfo{person}{Clemens Winter}, \bibinfo{person}{Chris Hesse},
  \bibinfo{person}{Mark Chen}, \bibinfo{person}{Eric Sigler},
  \bibinfo{person}{Mateusz Litwin}, \bibinfo{person}{Scott Gray},
  \bibinfo{person}{Benjamin Chess}, \bibinfo{person}{Jack Clark},
  \bibinfo{person}{Christopher Berner}, \bibinfo{person}{Sam McCandlish},
  \bibinfo{person}{Alec Radford}, \bibinfo{person}{Ilya Sutskever}, {and}
  \bibinfo{person}{Dario Amodei}.} \bibinfo{year}{2020}\natexlab{}.
\newblock \showarticletitle{Language Models are Few-Shot Learners}. In
  \bibinfo{booktitle}{\emph{Advances in Neural Information Processing
  Systems}}, \bibfield{editor}{\bibinfo{person}{H.~Larochelle},
  \bibinfo{person}{M.~Ranzato}, \bibinfo{person}{R.~Hadsell},
  \bibinfo{person}{M.~F. Balcan}, {and} \bibinfo{person}{H.~Lin}} (Eds.),
  Vol.~\bibinfo{volume}{33}. \bibinfo{publisher}{Curran Associates, Inc.},
  \bibinfo{pages}{1877--1901}.
\newblock
\urldef\tempurl%
\url{https://proceedings.neurips.cc/paper/2020/file/1457c0d6bfcb4967418bfb8ac142f64a-Paper.pdf}
\showURL{%
\tempurl}


\bibitem[\protect\citeauthoryear{Chen, Tworek, Jun, Yuan, de~Oliveira~Pinto,
  Kaplan, Edwards, Burda, Joseph, Brockman, Ray, Puri, Krueger, Petrov, Khlaaf,
  Sastry, Mishkin, Chan, Gray, Ryder, Pavlov, Power, Kaiser, Bavarian, Winter,
  Tillet, Such, Cummings, Plappert, Chantzis, Barnes, Herbert{-}Voss, Guss,
  Nichol, Paino, Tezak, Tang, Babuschkin, Balaji, Jain, Saunders, Hesse, Carr,
  Leike, Achiam, Misra, Morikawa, Radford, Knight, Brundage, Murati, Mayer,
  Welinder, McGrew, Amodei, McCandlish, Sutskever, and Zaremba}{Chen
  et~al\mbox{.}}{2021}]%
        {DBLP:journals/corr/abs-2107-03374}
\bibfield{author}{\bibinfo{person}{Mark Chen}, \bibinfo{person}{Jerry Tworek},
  \bibinfo{person}{Heewoo Jun}, \bibinfo{person}{Qiming Yuan},
  \bibinfo{person}{Henrique~Ponde de Oliveira~Pinto}, \bibinfo{person}{Jared
  Kaplan}, \bibinfo{person}{Harrison Edwards}, \bibinfo{person}{Yuri Burda},
  \bibinfo{person}{Nicholas Joseph}, \bibinfo{person}{Greg Brockman},
  \bibinfo{person}{Alex Ray}, \bibinfo{person}{Raul Puri},
  \bibinfo{person}{Gretchen Krueger}, \bibinfo{person}{Michael Petrov},
  \bibinfo{person}{Heidy Khlaaf}, \bibinfo{person}{Girish Sastry},
  \bibinfo{person}{Pamela Mishkin}, \bibinfo{person}{Brooke Chan},
  \bibinfo{person}{Scott Gray}, \bibinfo{person}{Nick Ryder},
  \bibinfo{person}{Mikhail Pavlov}, \bibinfo{person}{Alethea Power},
  \bibinfo{person}{Lukasz Kaiser}, \bibinfo{person}{Mohammad Bavarian},
  \bibinfo{person}{Clemens Winter}, \bibinfo{person}{Philippe Tillet},
  \bibinfo{person}{Felipe~Petroski Such}, \bibinfo{person}{Dave Cummings},
  \bibinfo{person}{Matthias Plappert}, \bibinfo{person}{Fotios Chantzis},
  \bibinfo{person}{Elizabeth Barnes}, \bibinfo{person}{Ariel Herbert{-}Voss},
  \bibinfo{person}{William~Hebgen Guss}, \bibinfo{person}{Alex Nichol},
  \bibinfo{person}{Alex Paino}, \bibinfo{person}{Nikolas Tezak},
  \bibinfo{person}{Jie Tang}, \bibinfo{person}{Igor Babuschkin},
  \bibinfo{person}{Suchir Balaji}, \bibinfo{person}{Shantanu Jain},
  \bibinfo{person}{William Saunders}, \bibinfo{person}{Christopher Hesse},
  \bibinfo{person}{Andrew~N. Carr}, \bibinfo{person}{Jan Leike},
  \bibinfo{person}{Joshua Achiam}, \bibinfo{person}{Vedant Misra},
  \bibinfo{person}{Evan Morikawa}, \bibinfo{person}{Alec Radford},
  \bibinfo{person}{Matthew Knight}, \bibinfo{person}{Miles Brundage},
  \bibinfo{person}{Mira Murati}, \bibinfo{person}{Katie Mayer},
  \bibinfo{person}{Peter Welinder}, \bibinfo{person}{Bob McGrew},
  \bibinfo{person}{Dario Amodei}, \bibinfo{person}{Sam McCandlish},
  \bibinfo{person}{Ilya Sutskever}, {and} \bibinfo{person}{Wojciech Zaremba}.}
  \bibinfo{year}{2021}\natexlab{}.
\newblock \showarticletitle{Evaluating Large Language Models Trained on Code}.
\newblock \bibinfo{journal}{\emph{CoRR}}  \bibinfo{volume}{abs/2107.03374}
  (\bibinfo{year}{2021}).
\newblock
\showeprint[arxiv]{2107.03374}
\urldef\tempurl%
\url{https://arxiv.org/abs/2107.03374}
\showURL{%
\tempurl}


\bibitem[\protect\citeauthoryear{Chen, Wang, Ye, Durrett, and Dillig}{Chen
  et~al\mbox{.}}{2020}]%
        {regel}
\bibfield{author}{\bibinfo{person}{Qiaochu Chen}, \bibinfo{person}{Xinyu Wang},
  \bibinfo{person}{Xi Ye}, \bibinfo{person}{Greg Durrett}, {and}
  \bibinfo{person}{Isil Dillig}.} \bibinfo{year}{2020}\natexlab{}.
\newblock \showarticletitle{Multi-Modal Synthesis of Regular Expressions}. In
  \bibinfo{booktitle}{\emph{Proceedings of the 41st ACM SIGPLAN Conference on
  Programming Language Design and Implementation}} \emph{(\bibinfo{series}{PLDI
  2020})}. \bibinfo{publisher}{Association for Computing Machinery},
  \bibinfo{address}{New York, NY, USA}, \bibinfo{pages}{487–502}.
\newblock
\showISBNx{9781450376136}
\urldef\tempurl%
\url{https://doi.org/10.1145/3385412.3385988}
\showDOI{\tempurl}


\bibitem[\protect\citeauthoryear{Chen, Martins, and Feng}{Chen
  et~al\mbox{.}}{2019}]%
        {chen2019maximal}
\bibfield{author}{\bibinfo{person}{Yanju Chen}, \bibinfo{person}{Ruben
  Martins}, {and} \bibinfo{person}{Yu Feng}.} \bibinfo{year}{2019}\natexlab{}.
\newblock \showarticletitle{Maximal multi-layer specification synthesis}. In
  \bibinfo{booktitle}{\emph{Proceedings of the 2019 27th ACM Joint Meeting on
  European Software Engineering Conference and Symposium on the Foundations of
  Software Engineering}}. \bibinfo{pages}{602--612}.
\newblock


\bibitem[\protect\citeauthoryear{Devlin, Chang, Lee, and Toutanova}{Devlin
  et~al\mbox{.}}{2019}]%
        {bert}
\bibfield{author}{\bibinfo{person}{Jacob Devlin}, \bibinfo{person}{Ming-Wei
  Chang}, \bibinfo{person}{Kenton Lee}, {and} \bibinfo{person}{Kristina
  Toutanova}.} \bibinfo{year}{2019}\natexlab{}.
\newblock \showarticletitle{{BERT}: Pre-training of Deep Bidirectional
  Transformers for Language Understanding}. In
  \bibinfo{booktitle}{\emph{Proceedings of the 2019 Conference of the North
  {A}merican Chapter of the Association for Computational Linguistics: Human
  Language Technologies, Volume 1 (Long and Short Papers)}}.
  \bibinfo{publisher}{Association for Computational Linguistics},
  \bibinfo{address}{Minneapolis, Minnesota}, \bibinfo{pages}{4171--4186}.
\newblock
\urldef\tempurl%
\url{https://doi.org/10.18653/v1/N19-1423}
\showDOI{\tempurl}


\bibitem[\protect\citeauthoryear{Drachsler-Cohen, Shoham, and
  Yahav}{Drachsler-Cohen et~al\mbox{.}}{2017}]%
        {abstract-synthesis}
\bibfield{author}{\bibinfo{person}{Dana Drachsler-Cohen},
  \bibinfo{person}{Sharon Shoham}, {and} \bibinfo{person}{Eran Yahav}.}
  \bibinfo{year}{2017}\natexlab{}.
\newblock \showarticletitle{Synthesis with Abstract Examples}. In
  \bibinfo{booktitle}{\emph{Computer Aided Verification}},
  \bibfield{editor}{\bibinfo{person}{Rupak Majumdar} {and}
  \bibinfo{person}{Viktor Kun{\v{c}}ak}} (Eds.). \bibinfo{publisher}{Springer
  International Publishing}, \bibinfo{address}{Cham},
  \bibinfo{pages}{254--278}.
\newblock
\showISBNx{978-3-319-63387-9}


\bibitem[\protect\citeauthoryear{Feng, Martins, Van~Geffen, Dillig, and
  Chaudhuri}{Feng et~al\mbox{.}}{2017a}]%
        {component-table}
\bibfield{author}{\bibinfo{person}{Yu Feng}, \bibinfo{person}{Ruben Martins},
  \bibinfo{person}{Jacob Van~Geffen}, \bibinfo{person}{Isil Dillig}, {and}
  \bibinfo{person}{Swarat Chaudhuri}.} \bibinfo{year}{2017}\natexlab{a}.
\newblock \showarticletitle{Component-Based Synthesis of Table Consolidation
  and Transformation Tasks from Examples}. In
  \bibinfo{booktitle}{\emph{Proceedings of the 38th ACM SIGPLAN Conference on
  Programming Language Design and Implementation}} \emph{(\bibinfo{series}{PLDI
  2017})}. \bibinfo{publisher}{Association for Computing Machinery},
  \bibinfo{address}{New York, NY, USA}, \bibinfo{pages}{422–436}.
\newblock
\showISBNx{9781450349888}
\urldef\tempurl%
\url{https://doi.org/10.1145/3062341.3062351}
\showDOI{\tempurl}


\bibitem[\protect\citeauthoryear{Feng, Martins, Wang, Dillig, and Reps}{Feng
  et~al\mbox{.}}{2017b}]%
        {Feng2017}
\bibfield{author}{\bibinfo{person}{Yu Feng}, \bibinfo{person}{Ruben Martins},
  \bibinfo{person}{Yuepeng Wang}, \bibinfo{person}{Isil Dillig}, {and}
  \bibinfo{person}{Thomas~W. Reps}.} \bibinfo{year}{2017}\natexlab{b}.
\newblock \showarticletitle{Component-based synthesis for complex {APIs}}. In
  \bibinfo{booktitle}{\emph{Proceedings of the 44th {ACM} {SIGPLAN} Symposium
  on Principles of Programming Languages}}. \bibinfo{publisher}{{ACM}}.
\newblock
\urldef\tempurl%
\url{https://doi.org/10.1145/3009837.3009851}
\showDOI{\tempurl}


\bibitem[\protect\citeauthoryear{Gulwani}{Gulwani}{2011}]%
        {flashfill}
\bibfield{author}{\bibinfo{person}{Sumit Gulwani}.}
  \bibinfo{year}{2011}\natexlab{}.
\newblock \showarticletitle{Automating String Processing in Spreadsheets using
  Input-Output Examples}. In \bibinfo{booktitle}{\emph{PoPL'11, January 26-28,
  2011, Austin, Texas, USA}}.
\newblock
\urldef\tempurl%
\url{https://www.microsoft.com/en-us/research/publication/automating-string-processing-spreadsheets-using-input-output-examples/}
\showURL{%
\tempurl}


\bibitem[\protect\citeauthoryear{Gulwani, Jha, Tiwari, and Venkatesan}{Gulwani
  et~al\mbox{.}}{2011}]%
        {gulwani2011loopfree}
\bibfield{author}{\bibinfo{person}{Sumit Gulwani}, \bibinfo{person}{Susmit
  Jha}, \bibinfo{person}{Ashish Tiwari}, {and} \bibinfo{person}{Ramarathnam
  Venkatesan}.} \bibinfo{year}{2011}\natexlab{}.
\newblock \showarticletitle{Synthesis of Loop-Free Programs}. In
  \bibinfo{booktitle}{\emph{Proceedings of the 32nd ACM SIGPLAN Conference on
  Programming Language Design and Implementation}} \emph{(\bibinfo{series}{PLDI
  '11})}. \bibinfo{publisher}{Association for Computing Machinery},
  \bibinfo{address}{New York, NY, USA}, \bibinfo{pages}{62–73}.
\newblock
\showISBNx{9781450306638}
\urldef\tempurl%
\url{https://doi.org/10.1145/1993498.1993506}
\showDOI{\tempurl}


\bibitem[\protect\citeauthoryear{Gulwani and Marron}{Gulwani and
  Marron}{2014}]%
        {nlyze}
\bibfield{author}{\bibinfo{person}{Sumit Gulwani} {and} \bibinfo{person}{Mark
  Marron}.} \bibinfo{year}{2014}\natexlab{}.
\newblock \showarticletitle{NLyze: Interactive Programming by Natural Language
  for SpreadSheet Data Analysis and Manipulation}. In
  \bibinfo{booktitle}{\emph{SIGMOD '14 Proceedings of the 2014 ACM SIGMOD
  International Conference on Management of Data} (\bibinfo{edition}{sigmod '14
  proceedings of the 2014 acm sigmod international conference on management of
  data} ed.)}. \bibinfo{publisher}{Association for Computing Machinery},
  \bibinfo{pages}{803--814}.
\newblock
\urldef\tempurl%
\url{https://www.microsoft.com/en-us/research/publication/nlyze-interactive-programming-by-natural-language-for-spreadsheet-data-analysis-and-manipulation/}
\showURL{%
\tempurl}


\bibitem[\protect\citeauthoryear{Hendrycks, Basart, Kadavath, Mazeika, Arora,
  Guo, Burns, Puranik, He, Song, and Steinhardt}{Hendrycks
  et~al\mbox{.}}{2021}]%
        {DBLP:journals/corr/abs-2105-09938}
\bibfield{author}{\bibinfo{person}{Dan Hendrycks}, \bibinfo{person}{Steven
  Basart}, \bibinfo{person}{Saurav Kadavath}, \bibinfo{person}{Mantas Mazeika},
  \bibinfo{person}{Akul Arora}, \bibinfo{person}{Ethan Guo},
  \bibinfo{person}{Collin Burns}, \bibinfo{person}{Samir Puranik},
  \bibinfo{person}{Horace He}, \bibinfo{person}{Dawn Song}, {and}
  \bibinfo{person}{Jacob Steinhardt}.} \bibinfo{year}{2021}\natexlab{}.
\newblock \showarticletitle{Measuring Coding Challenge Competence With {APPS}}.
\newblock \bibinfo{journal}{\emph{CoRR}}  \bibinfo{volume}{abs/2105.09938}
  (\bibinfo{year}{2021}).
\newblock
\showeprint[arxiv]{2105.09938}
\urldef\tempurl%
\url{https://arxiv.org/abs/2105.09938}
\showURL{%
\tempurl}


\bibitem[\protect\citeauthoryear{Huang, Qiu, Shen, and Wang}{Huang
  et~al\mbox{.}}{2020}]%
        {enumerative-deductive}
\bibfield{author}{\bibinfo{person}{Kangjing Huang}, \bibinfo{person}{Xiaokang
  Qiu}, \bibinfo{person}{Peiyuan Shen}, {and} \bibinfo{person}{Yanjun Wang}.}
  \bibinfo{year}{2020}\natexlab{}.
\newblock \showarticletitle{Reconciling Enumerative and Deductive Program
  Synthesis}. In \bibinfo{booktitle}{\emph{Proceedings of the 41st ACM SIGPLAN
  Conference on Programming Language Design and Implementation}}
  \emph{(\bibinfo{series}{PLDI 2020})}. \bibinfo{publisher}{Association for
  Computing Machinery}, \bibinfo{address}{New York, NY, USA},
  \bibinfo{pages}{1159–1174}.
\newblock
\showISBNx{9781450376136}
\urldef\tempurl%
\url{https://doi.org/10.1145/3385412.3386027}
\showDOI{\tempurl}


\bibitem[\protect\citeauthoryear{Huang, Wang, Singh, Yih, and He}{Huang
  et~al\mbox{.}}{2018}]%
        {meta-learning-natural}
\bibfield{author}{\bibinfo{person}{Po-Sen Huang}, \bibinfo{person}{Chenglong
  Wang}, \bibinfo{person}{Rishabh Singh}, \bibinfo{person}{Wen-tau Yih}, {and}
  \bibinfo{person}{Xiaodong He}.} \bibinfo{year}{2018}\natexlab{}.
\newblock \showarticletitle{Natural Language to Structured Query Generation via
  Meta-Learning}. In \bibinfo{booktitle}{\emph{Proceedings of the 2018
  Conference of the North {A}merican Chapter of the Association for
  Computational Linguistics: Human Language Technologies, Volume 2 (Short
  Papers)}}. \bibinfo{publisher}{Association for Computational Linguistics},
  \bibinfo{address}{New Orleans, Louisiana}, \bibinfo{pages}{732--738}.
\newblock
\urldef\tempurl%
\url{https://doi.org/10.18653/v1/N18-2115}
\showDOI{\tempurl}


\bibitem[\protect\citeauthoryear{Jha, Gulwani, Seshia, and Tiwari}{Jha
  et~al\mbox{.}}{2010}]%
        {oracle-guided}
\bibfield{author}{\bibinfo{person}{Susmit Jha}, \bibinfo{person}{Sumit
  Gulwani}, \bibinfo{person}{Sanjit~A. Seshia}, {and} \bibinfo{person}{Ashish
  Tiwari}.} \bibinfo{year}{2010}\natexlab{}.
\newblock \showarticletitle{Oracle-Guided Component-Based Program Synthesis}.
  In \bibinfo{booktitle}{\emph{ICSE '10, May 2-8 2010, Cape Town, South Africa}
  (\bibinfo{edition}{icse ’10, may 2-8 2010, cape town, south africa} ed.)}.
\newblock
\urldef\tempurl%
\url{https://www.microsoft.com/en-us/research/publication/oracle-guided-component-based-program-synthesis/}
\showURL{%
\tempurl}


\bibitem[\protect\citeauthoryear{Jones}{Jones}{1972}]%
        {sparckjones1972}
\bibfield{author}{\bibinfo{person}{K. Jones}.} \bibinfo{year}{1972}\natexlab{}.
\newblock \showarticletitle{A statistical interpretation of term specificity
  and its application in retrieval}.
\newblock \bibinfo{journal}{\emph{J. Documentation}}  \bibinfo{volume}{60}
  (\bibinfo{year}{1972}), \bibinfo{pages}{493--502}.
\newblock
\urldef\tempurl%
\url{https://doi.org/10.1108/eb026526}
\showDOI{\tempurl}


\bibitem[\protect\citeauthoryear{Kushman and Barzilay}{Kushman and
  Barzilay}{2013}]%
        {kushman2013}
\bibfield{author}{\bibinfo{person}{Nate Kushman} {and} \bibinfo{person}{Regina
  Barzilay}.} \bibinfo{year}{2013}\natexlab{}.
\newblock \showarticletitle{Using Semantic Unification to Generate Regular
  Expressions from Natural Language}. In \bibinfo{booktitle}{\emph{Proceedings
  of the 2013 Conference of the North {A}merican Chapter of the Association for
  Computational Linguistics: Human Language Technologies}}.
  \bibinfo{publisher}{Association for Computational Linguistics},
  \bibinfo{address}{Atlanta, Georgia}, \bibinfo{pages}{826--836}.
\newblock
\urldef\tempurl%
\url{https://www.aclweb.org/anthology/N13-1103}
\showURL{%
\tempurl}


\bibitem[\protect\citeauthoryear{Le, Gulwani, and Su}{Le et~al\mbox{.}}{2013}]%
        {smartsynth}
\bibfield{author}{\bibinfo{person}{Vu Le}, \bibinfo{person}{Sumit Gulwani},
  {and} \bibinfo{person}{Zhendong Su}.} \bibinfo{year}{2013}\natexlab{}.
\newblock \showarticletitle{SmartSynth: Synthesizing Smartphone Automation
  Scripts from Natural Language}. In \bibinfo{booktitle}{\emph{MobiSys'13, June
  25-28, 2013, Taipei, Taiwan} (\bibinfo{edition}{mobisys’13, june 25-28,
  2013, taipei, taiwan} ed.)}.
\newblock
\urldef\tempurl%
\url{https://www.microsoft.com/en-us/research/publication/smartsynth-synthesizing-smartphone-automation-scripts-natural-language/}
\showURL{%
\tempurl}


\bibitem[\protect\citeauthoryear{Lee, So, and Oh}{Lee et~al\mbox{.}}{2016}]%
        {mina2017}
\bibfield{author}{\bibinfo{person}{Mina Lee}, \bibinfo{person}{Sunbeom So},
  {and} \bibinfo{person}{Hakjoo Oh}.} \bibinfo{year}{2016}\natexlab{}.
\newblock \showarticletitle{Synthesizing Regular Expressions from Examples for
  Introductory Automata Assignments}.
\newblock \bibinfo{journal}{\emph{SIGPLAN Not.}} \bibinfo{volume}{52},
  \bibinfo{number}{3} (\bibinfo{date}{Oct.} \bibinfo{year}{2016}),
  \bibinfo{pages}{70–80}.
\newblock
\showISSN{0362-1340}
\urldef\tempurl%
\url{https://doi.org/10.1145/3093335.2993244}
\showDOI{\tempurl}


\bibitem[\protect\citeauthoryear{Li, Li, Xu, Cao, Chen, Hu, Chen, and
  Cheung}{Li et~al\mbox{.}}{2020}]%
        {DBLP:journals/corr/abs-2012-15489}
\bibfield{author}{\bibinfo{person}{Yeting Li}, \bibinfo{person}{Shuaimin Li},
  \bibinfo{person}{Zhiwu Xu}, \bibinfo{person}{Jialun Cao},
  \bibinfo{person}{Zixuan Chen}, \bibinfo{person}{Yun Hu},
  \bibinfo{person}{Haiming Chen}, {and} \bibinfo{person}{Shing{-}Chi Cheung}.}
  \bibinfo{year}{2020}\natexlab{}.
\newblock \showarticletitle{TransRegex: Multi-modal Regular Expression
  Synthesis by Generate-and-Repair}.
\newblock \bibinfo{journal}{\emph{CoRR}}  \bibinfo{volume}{abs/2012.15489}
  (\bibinfo{year}{2020}).
\newblock
\showeprint[arxiv]{2012.15489}
\urldef\tempurl%
\url{https://arxiv.org/abs/2012.15489}
\showURL{%
\tempurl}


\bibitem[\protect\citeauthoryear{Lin, Wang, Zettlemoyer, and Ernst}{Lin
  et~al\mbox{.}}{2018}]%
        {nl2bash}
\bibfield{author}{\bibinfo{person}{Xi~Victoria Lin}, \bibinfo{person}{Chenglong
  Wang}, \bibinfo{person}{Luke Zettlemoyer}, {and} \bibinfo{person}{Michael~D.
  Ernst}.} \bibinfo{year}{2018}\natexlab{}.
\newblock \showarticletitle{{NL}2{B}ash: A Corpus and Semantic Parser for
  Natural Language Interface to the Linux Operating System}. In
  \bibinfo{booktitle}{\emph{Proceedings of the Eleventh International
  Conference on Language Resources and Evaluation ({LREC} 2018)}}.
  \bibinfo{publisher}{European Language Resources Association (ELRA)},
  \bibinfo{address}{Miyazaki, Japan}.
\newblock
\urldef\tempurl%
\url{https://www.aclweb.org/anthology/L18-1491}
\showURL{%
\tempurl}


\bibitem[\protect\citeauthoryear{Locascio, Narasimhan, DeLeon, Kushman, and
  Barzilay}{Locascio et~al\mbox{.}}{2016}]%
        {deepregex}
\bibfield{author}{\bibinfo{person}{Nicholas Locascio}, \bibinfo{person}{Karthik
  Narasimhan}, \bibinfo{person}{Eduardo DeLeon}, \bibinfo{person}{Nate
  Kushman}, {and} \bibinfo{person}{Regina Barzilay}.}
  \bibinfo{year}{2016}\natexlab{}.
\newblock \showarticletitle{Neural Generation of Regular Expressions from
  Natural Language with Minimal Domain Knowledge}. In
  \bibinfo{booktitle}{\emph{EMNLP} (\bibinfo{edition}{emnlp} ed.)}.
\newblock
\urldef\tempurl%
\url{https://www.microsoft.com/en-us/research/publication/neural-generation-regular-expressions-natural-language-minimal-domain-knowledge/}
\showURL{%
\tempurl}


\bibitem[\protect\citeauthoryear{Manshadi, Gildea, and Allen}{Manshadi
  et~al\mbox{.}}{2013}]%
        {manshadi2013}
\bibfield{author}{\bibinfo{person}{Mehdi Manshadi}, \bibinfo{person}{Daniel
  Gildea}, {and} \bibinfo{person}{James Allen}.}
  \bibinfo{year}{2013}\natexlab{}.
\newblock \showarticletitle{Integrating programming by example and natural
  language programming}. In \bibinfo{booktitle}{\emph{Proceedings of the AAAI
  Conference on Artificial Intelligence}}, Vol.~\bibinfo{volume}{27}.
\newblock


\bibitem[\protect\citeauthoryear{Meyes, Lu, de~Puiseau, and Meisen}{Meyes
  et~al\mbox{.}}{2019}]%
        {DBLP:journals/corr/abs-1901-08644}
\bibfield{author}{\bibinfo{person}{Richard Meyes}, \bibinfo{person}{Melanie
  Lu}, \bibinfo{person}{Constantin~Waubert de Puiseau}, {and}
  \bibinfo{person}{Tobias Meisen}.} \bibinfo{year}{2019}\natexlab{}.
\newblock \showarticletitle{Ablation Studies in Artificial Neural Networks}.
\newblock \bibinfo{journal}{\emph{CoRR}}  \bibinfo{volume}{abs/1901.08644}
  (\bibinfo{year}{2019}).
\newblock
\showeprint[arxiv]{1901.08644}
\urldef\tempurl%
\url{http://arxiv.org/abs/1901.08644}
\showURL{%
\tempurl}


\bibitem[\protect\citeauthoryear{OpenAI}{OpenAI}{2021}]%
        {gpt3apps}
\bibfield{author}{\bibinfo{person}{OpenAI}.} \bibinfo{year}{2021}\natexlab{}.
\newblock \bibinfo{title}{GPT-3 powers the next generation of apps}.
\newblock
\newblock
\newblock
\shownote{\url{https://openai.com/blog/gpt-3-apps/}.}


\bibitem[\protect\citeauthoryear{Pan, Hu, Xu, and D'Antoni}{Pan
  et~al\mbox{.}}{2019}]%
        {rfixer}
\bibfield{author}{\bibinfo{person}{Rong Pan}, \bibinfo{person}{Qinheping Hu},
  \bibinfo{person}{Gaowei Xu}, {and} \bibinfo{person}{Loris D'Antoni}.}
  \bibinfo{year}{2019}\natexlab{}.
\newblock \showarticletitle{Automatic Repair of Regular Expressions}.
\newblock \bibinfo{journal}{\emph{Proc. ACM Program. Lang.}}
  \bibinfo{volume}{3}, \bibinfo{number}{OOPSLA}, Article
  \bibinfo{articleno}{139} (\bibinfo{date}{Oct.} \bibinfo{year}{2019}),
  \bibinfo{numpages}{29}~pages.
\newblock
\urldef\tempurl%
\url{https://doi.org/10.1145/3360565}
\showDOI{\tempurl}


\bibitem[\protect\citeauthoryear{Pei, Han, and Mao}{Pei et~al\mbox{.}}{2000}]%
        {phm00}
\bibfield{author}{\bibinfo{person}{Jian Pei}, \bibinfo{person}{Jiawei Han},
  {and} \bibinfo{person}{Runying Mao}.} \bibinfo{year}{2000}\natexlab{}.
\newblock \showarticletitle{CLOSET: An Efficient Algorithm for Mining Frequent
  Closed Itemsets.}. In \bibinfo{booktitle}{\emph{ACM SIGMOD Workshop on
  Research Issues in Data Mining and Knowledge Discovery}} (2002-01-03).
  \bibinfo{pages}{21--30}.
\newblock
\urldef\tempurl%
\url{http://dblp.uni-trier.de/db/conf/dmkd/dmkd2000.html#PeiHM00}
\showURL{%
\tempurl}


\bibitem[\protect\citeauthoryear{Polikarpova, Kuraj, and
  Solar-Lezama}{Polikarpova et~al\mbox{.}}{2016}]%
        {refinement-type}
\bibfield{author}{\bibinfo{person}{Nadia Polikarpova}, \bibinfo{person}{Ivan
  Kuraj}, {and} \bibinfo{person}{Armando Solar-Lezama}.}
  \bibinfo{year}{2016}\natexlab{}.
\newblock \showarticletitle{Program Synthesis from Polymorphic Refinement
  Types}.
\newblock \bibinfo{journal}{\emph{SIGPLAN Not.}} \bibinfo{volume}{51},
  \bibinfo{number}{6} (\bibinfo{date}{June} \bibinfo{year}{2016}),
  \bibinfo{pages}{522–538}.
\newblock
\showISSN{0362-1340}
\urldef\tempurl%
\url{https://doi.org/10.1145/2980983.2908093}
\showDOI{\tempurl}


\bibitem[\protect\citeauthoryear{Raza and Gulwani}{Raza and Gulwani}{2017}]%
        {raza2017automated}
\bibfield{author}{\bibinfo{person}{Mohammad Raza} {and} \bibinfo{person}{Sumit
  Gulwani}.} \bibinfo{year}{2017}\natexlab{}.
\newblock \showarticletitle{Automated Data Extraction Using Predictive Program
  Synthesis}. In \bibinfo{booktitle}{\emph{Proceedings of the Thirty-First AAAI
  Conference on Artificial Intelligence}} \emph{(\bibinfo{series}{AAAI'17})}.
  \bibinfo{publisher}{AAAI Press}, \bibinfo{pages}{882–890}.
\newblock


\bibitem[\protect\citeauthoryear{Raza and Gulwani}{Raza and Gulwani}{2020}]%
        {raza2020web}
\bibfield{author}{\bibinfo{person}{Mohammad Raza} {and} \bibinfo{person}{Sumit
  Gulwani}.} \bibinfo{year}{2020}\natexlab{}.
\newblock \showarticletitle{Web Data Extraction Using Hybrid Program Synthesis:
  A Combination of Top-down and Bottom-up Inference}. In
  \bibinfo{booktitle}{\emph{Proceedings of the 2020 ACM SIGMOD International
  Conference on Management of Data}} \emph{(\bibinfo{series}{SIGMOD '20})}.
  \bibinfo{publisher}{Association for Computing Machinery},
  \bibinfo{address}{New York, NY, USA}, \bibinfo{pages}{1967–1978}.
\newblock
\showISBNx{9781450367356}
\urldef\tempurl%
\url{https://doi.org/10.1145/3318464.3380608}
\showDOI{\tempurl}


\bibitem[\protect\citeauthoryear{Raza, Gulwani, and Milic-Frayling}{Raza
  et~al\mbox{.}}{2015}]%
        {raza2015}
\bibfield{author}{\bibinfo{person}{Mohammad Raza}, \bibinfo{person}{Sumit
  Gulwani}, {and} \bibinfo{person}{Natasa Milic-Frayling}.}
  \bibinfo{year}{2015}\natexlab{}.
\newblock \showarticletitle{Compositional Program Synthesis from Natural
  Language and Examples}. In \bibinfo{booktitle}{\emph{IJCAI 2015}
  (\bibinfo{edition}{ijcai 2015} ed.)}.
\newblock
\urldef\tempurl%
\url{https://www.microsoft.com/en-us/research/publication/compositional-program-synthesis-natural-language-examples/}
\showURL{%
\tempurl}


\bibitem[\protect\citeauthoryear{Solar-Lezama}{Solar-Lezama}{2008}]%
        {sketching}
\bibfield{author}{\bibinfo{person}{Armando Solar-Lezama}.}
  \bibinfo{year}{2008}\natexlab{}.
\newblock \emph{\bibinfo{title}{Program Synthesis by Sketching}}.
\newblock \bibinfo{thesistype}{Ph.D. Dissertation}. \bibinfo{address}{USA}.
\newblock Advisor(s) Bodik, Rastislav.
\newblock
\showISBNx{9781109097450}
\newblock
\shownote{AAI3353225.}


\bibitem[\protect\citeauthoryear{Srivastava, Gulwani, and Foster}{Srivastava
  et~al\mbox{.}}{2010}]%
        {verification-synthesis}
\bibfield{author}{\bibinfo{person}{Saurabh Srivastava}, \bibinfo{person}{Sumit
  Gulwani}, {and} \bibinfo{person}{Jeffrey~S. Foster}.}
  \bibinfo{year}{2010}\natexlab{}.
\newblock \showarticletitle{From Program Verification to Program Synthesis}. In
  \bibinfo{booktitle}{\emph{Proceedings of the 37th Annual ACM SIGPLAN-SIGACT
  Symposium on Principles of Programming Languages}}
  \emph{(\bibinfo{series}{POPL '10})}. \bibinfo{publisher}{Association for
  Computing Machinery}, \bibinfo{address}{New York, NY, USA},
  \bibinfo{pages}{313–326}.
\newblock
\showISBNx{9781605584799}
\urldef\tempurl%
\url{https://doi.org/10.1145/1706299.1706337}
\showDOI{\tempurl}


\bibitem[\protect\citeauthoryear{W3C}{W3C}{2020}]%
        {css}
\bibfield{author}{\bibinfo{person}{W3C}.} \bibinfo{year}{2020}\natexlab{}.
\newblock \bibinfo{title}{CSS Snapshot 2020 [online]}.
\newblock
\newblock
\urldef\tempurl%
\url{https://www.w3.org/TR/CSS/}
\showURL{%
\tempurl}


\bibitem[\protect\citeauthoryear{Wang, Shin, Liu, Polozov, and Richardson}{Wang
  et~al\mbox{.}}{2020}]%
        {rat-sql}
\bibfield{author}{\bibinfo{person}{Bailin Wang}, \bibinfo{person}{Richard
  Shin}, \bibinfo{person}{Xiaodong Liu}, \bibinfo{person}{Oleksandr Polozov},
  {and} \bibinfo{person}{Matthew Richardson}.} \bibinfo{year}{2020}\natexlab{}.
\newblock \showarticletitle{{RAT-SQL}: Relation-Aware Schema Encoding and
  Linking for Text-to-{SQL} Parsers}. In \bibinfo{booktitle}{\emph{Proceedings
  of the 58th Annual Meeting of the Association for Computational
  Linguistics}}. \bibinfo{publisher}{Association for Computational
  Linguistics}, \bibinfo{address}{Online}, \bibinfo{pages}{7567--7578}.
\newblock
\urldef\tempurl%
\url{https://doi.org/10.18653/v1/2020.acl-main.677}
\showDOI{\tempurl}


\bibitem[\protect\citeauthoryear{Yaghmazadeh, Wang, Dillig, and
  Dillig}{Yaghmazadeh et~al\mbox{.}}{2017}]%
        {sqlizer}
\bibfield{author}{\bibinfo{person}{Navid Yaghmazadeh}, \bibinfo{person}{Yuepeng
  Wang}, \bibinfo{person}{Isil Dillig}, {and} \bibinfo{person}{Thomas Dillig}.}
  \bibinfo{year}{2017}\natexlab{}.
\newblock \showarticletitle{SQLizer: Query Synthesis from Natural Language}.
\newblock \bibinfo{journal}{\emph{Proc. ACM Program. Lang.}}
  \bibinfo{volume}{1}, \bibinfo{number}{OOPSLA}, Article
  \bibinfo{articleno}{63} (\bibinfo{date}{Oct.} \bibinfo{year}{2017}),
  \bibinfo{numpages}{26}~pages.
\newblock
\urldef\tempurl%
\url{https://doi.org/10.1145/3133887}
\showDOI{\tempurl}


\bibitem[\protect\citeauthoryear{Zhong, Guo, Yang, Peng, Xie, Lou, Liu, and
  Zhang}{Zhong et~al\mbox{.}}{2018}]%
        {semregex}
\bibfield{author}{\bibinfo{person}{Zexuan Zhong}, \bibinfo{person}{Jiaqi Guo},
  \bibinfo{person}{Wei Yang}, \bibinfo{person}{Jian Peng}, \bibinfo{person}{Tao
  Xie}, \bibinfo{person}{Jian-Guang Lou}, \bibinfo{person}{Ting Liu}, {and}
  \bibinfo{person}{Dongmei Zhang}.} \bibinfo{year}{2018}\natexlab{}.
\newblock \showarticletitle{SemRegex: A Semantics-Based Approach for Generating
  Regular Expressions from Natural Language Specifications}. In
  \bibinfo{booktitle}{\emph{EMNP'18}}. \bibinfo{publisher}{ACL}.
\newblock
\urldef\tempurl%
\url{https://www.microsoft.com/en-us/research/publication/semregex-a-semantics-based-approach-for-generating-regular-expressions-from-natural-language-specifications-2/}
\showURL{%
\tempurl}


\end{thebibliography}
